\documentclass[sigconf]{acmart}

\makeatletter
\def\@ACM@checkaffil{% Only warnings
    \if@ACM@instpresent\else
    \ClassWarningNoLine{\@classname}{No institution present for an affiliation}%
    \fi
    \if@ACM@citypresent\else
    \ClassWarningNoLine{\@classname}{No city present for an affiliation}%
    \fi
    \if@ACM@countrypresent\else
        \ClassWarningNoLine{\@classname}{No country present for an affiliation}%
    \fi
}
\makeatother
\AtBeginDocument{%
  \providecommand\BibTeX{{%
    \normalfont B\kern-0.5em{\scshape i\kern-0.25em b}\kern-0.8em\TeX}}}

\usepackage{multirow}
\usepackage{subcaption}
\usepackage{comment}
\usepackage{algorithm}
\usepackage{algpseudocode}

\usepackage[capitalize]{cleveref}
\crefname{section}{Sec.}{Secs.}
\Crefname{section}{Section}{Sections}
\Crefname{table}{Table}{Tables}
\crefname{table}{Tab.}{Tabs.}

\usepackage{xspace}
\makeatletter
\DeclareRobustCommand\onedot{\futurelet\@let@token\@onedot}
\def\@onedot{\ifx\@let@token.\else.\null\fi\xspace}

\def\ie{\emph{i.e}\onedot}

\def\wrt{w.r.t\onedot} 
 
\def\etal{\emph{et al}\onedot}
\makeatother

\setcopyright{acmcopyright}
\copyrightyear{2022}
\acmYear{2022}
\acmDOI{10.1145/3551626.3564942}

\acmConference[MMAsia '22]{ACM Multimedia Asia}{December 13--16, 2022}{Tokyo, Japan}
\acmBooktitle{ACM Multimedia Asia (MMAsia '22), December 13--16, 2022, Tokyo, Japan}
\acmPrice{15.00}
\acmISBN{978-1-4503-9478-9/22/12}

\acmSubmissionID{39}

\begin{document}

\title{Informative Sample-Aware Proxy for Deep Metric Learning}


\author{Aoyu Li}
 \email{aoyuli@rio.gsic.titech.ac.jp}
\affiliation{
  \institution{Tokyo Institute of Technology}
  %\city{Tokyo}
%  \country{Japan}
}

\author{Ikuro Sato}
%\email{isato@c.titech.ac.jp}
\affiliation{
  \institution{Tokyo Institute of Technology}
  \institution{Denso IT Laboratory}
%  \city{Tokyo}
%  \country{Japan}
}

\author{Kohta Ishikawa}
%\email{ishikawa.kohta@core.d-itlab.co.jp}
\affiliation{%
  \institution{Denso IT Laboratory}
%  \city{Tokyo}
%  \country{Japan}
}

\author{Rei Kawakami}
%\email{reikawa@sc.e.titech.ac.jp}
\affiliation{
  \institution{Tokyo Institute of Technology}
%  \city{Tokyo}
%  \country{Japan}
}

\author{Rio Yokota}
%\email{rioyokota@gsic.titech.ac.jp}
\affiliation{%
  \institution{Tokyo Institute of Technology}
%   \streetaddress{}
%  \city{Tokyo}
%   \state{}
%  \country{Japan}
%   \postcode{}
}

\renewcommand{\shortauthors}{Li, et al.}
% \renewcommand{\shortauthors}{Anonymized author, et al.}

\begin{abstract}
    Among various supervised deep metric learning methods proxy-based approaches have achieved high retrieval accuracies.
    Proxies, which are class-representative points in an embedding space, receive updates based on proxy-sample similarities in a similar manner to sample representations.
    In existing methods, a relatively small number of samples can produce large gradient magnitudes (\ie, hard samples), and a relatively large number of samples can produce small gradient magnitudes (\ie, easy samples); these can play a major part in updates.
    Assuming that acquiring too much sensitivity to such extreme sets of samples would deteriorate the generalizability of a method, we propose a novel proxy-based method called Informative Sample-Aware Proxy (Proxy-ISA), which directly modifies a gradient weighting factor for each sample using a scheduled threshold function, so that the model is more sensitive to the informative samples.
    % Proxy-ISA reduces the factors to hard and easy samples using a scheduled threshold function so that the model is more sensitive to the intermediate samples.
    Extensive experiments on the CUB-200-2011, Cars-196, Stanford Online Products and In-shop Clothes Retrieval datasets demonstrate the superiority of Proxy-ISA compared with the state-of-the-art methods.
\end{abstract}



\begin{comment}
\begin{teaserfigure}
  \includegraphics[width=\textwidth]{}
  \caption{}
  \Description{}
  \label{fig:teaser}
\end{teaserfigure}
\end{comment}

\maketitle

\section{Introduction}
\label{sec:intro}

\begin{figure}[t]
    \vspace{2mm}
    \centering
    \includegraphics[width=\linewidth]{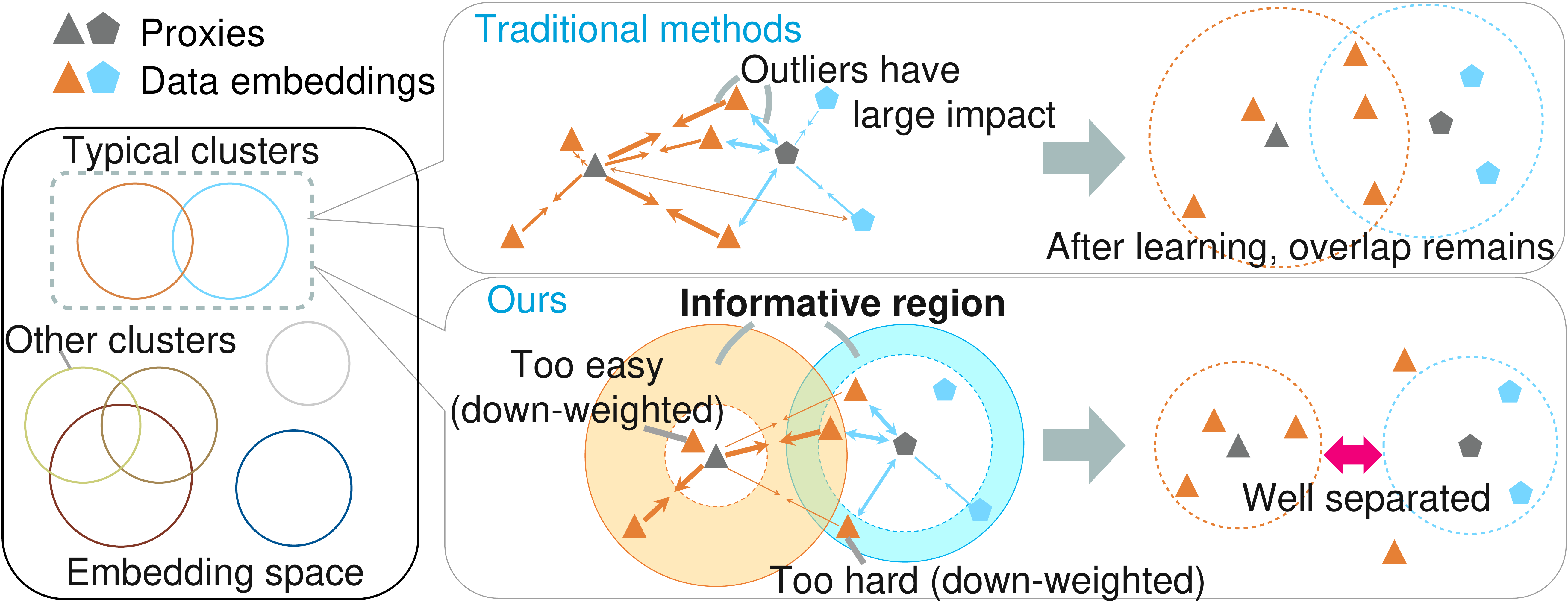}
    \caption{Overview of typical proxy-based DML and the proposed approach in this paper. Traditional methods can result in a few hard samples pulling the proxy, exacerbating the overlapping problem. The Proxy-ISA proposed in this paper aims to address this limitation by identifying such samples and reducing their contribution to the proxy.}
    \label{fig1.1}
    \vspace{-5mm}
\end{figure}

Learning semantic similarities between a pair of visual data is an important objective in computer vision tasks.
Metric learning aims to learn an embedding space such that samples from the same category are close to each other and otherwise far apart.
In deep metric learning (DML), learnable encoders are utilized to produce representations in an embedding space in which semantic distances between samples can be measured.
DML has broad computer-vision applications such as image retrieval \cite{NIPS2016_6b180037,RARS19}, face recognition \cite{DBLP:conf/eccv/WenZL016}, person re-identification \cite{WU2021107424} and few-shot learning \cite{DBLP:conf/cvpr/KarlinskySHSAFG19,DBLP:conf/cvpr/ZhaoLYC21,DBLP:journals/corr/abs-1909-09602}.

Throughout the past studies, DML methods can be categorized into two families: {\em pair-based} and {\em proxy-based}.
Pair-based methods consider the distance between a pair of data embeddings, while proxy-based methods consider the distance between a data embedding and a class-representative point in the embedding space.

A typical example of pair-based losses is contrastive loss \cite{1467314}, which aims to pull close or push apart a pair of data embeddings according to the identicality of the class labels. 
Other pair-based losses, such as triplet loss \cite{triplet} and N-pair loss \cite{NIPS2016_6b180037}, extend this idea to more than two samples.
Due to its combinatorial nature, a naive implementation of a pair-based method suffers from polynomial growth of complexity with respect to the number of training samples.
Additionally, training with pairs from a subset (mini-batch) lacks global information of the embedding space.
In contrast, proxy-based methods \cite{8237309,DBLP:conf/cvpr/WangWZJGZL018,Deng_2019_CVPR,qian2019striple} assign one or more trainable reference points called proxies to each class, and reduce the training complexity by considering only the distance from the data embeddings to proxies.
During training, proxies attract data embeddings of the same class (\ie, positives) and push away those of different classes (\ie, negatives).
Meanwhile, proxies are kept in memory and thus serve as a source of global information.

In existing methods, a small number of \textit{hard} samples, such as positives that are far apart and negatives that are close, produce large gradients, and thus will have more impact on learning.
A large number of \textit{easy} samples can also dominate the gradient because of their large population size.
An intuitive strategy is to select only the \textit{hard} samples \cite{DBLP:journals/corr/HermansBL17}, but this has been observed to produce noisy gradients and converge to bad local optima \cite{8237571}, and its impact on proxies remains to be discussed.
In pair-based methods, mining strategies, such as semi-hard negative mining \cite{Schroff_2015_CVPR}, have been proposed to select and learn more important negatives that are closer to the anchor but still farther away than the corresponding positives.
However, the \textit{hardness} is usually defined by a manual threshold, is fixed during learning, and is shared among different classes.
As in \cref{fig1.1}, increasing the gradient weights of hard proxy-data pairs may also disrupt the ideal distribution of the clusters.
% Typically, overlapping between clusters happens in the embedding space. Traditional proxy-based methods can cause the proxy to be pulled heavily by a few extremely hard samples, as shown in Figure 1. This may exacerbate the overlapping problem. On the other hand, the Proxy-ISA method proposed in this paper identifies such samples and reduces their contribution to the proxy, while also giving less attention to easier samples. This allows the proxy to focus on informative samples. The informative region is defined by two similarity thresholds to the proxy.

To address the limitations of existing proxy-based methods and to effectively incorporate the idea of sample importance into proxy-based DML, we propose the Informative Sample-Aware Proxy (Proxy-ISA) that directly controls gradient weights non-uniformly according to the temporal learning states, which are defined by how the samples with different hardness distribute in the embedding space.
Our idea is motivated by empirical findings in active learning \cite{DBLP:conf/nips/ChangLM17}, where informative samples dynamically change along learning.
In Proxy-ISA, as outlined in  \cref{fig1.1}, each proxy is assigned a temporal similarity range, determined by the space already learned (too easy), informative region, and the space contains outliers (too hard); weights for positive and negative pairs are assigned separately.
We apply a memory queue to estimate each class hardness.
Unlike memory-based DML \cite{wang2020xbm,ko2021learning}, where past information is directly used to update weights, our use of memory allows us to estimate the space occupied by easy samples around the proxies and to judge whether the sample of interest is informative.

Our main contributions are summarized as follows:
%\begin{itemize}
    \textbf{1)} We propose a novel method for proxy-based DML, which performs class-dependent dynamic weighting for each sample based on the learned intra- and inter-class relations.
    \textbf{2)} We apply the concept of a ``\textit{volume}" of the class-related region \cite{8953804} to DML and demonstrate empirically that focusing on informative samples based on the estimation of the class hardness improves generalization.
    \textbf{3)} Proxy-ISA achieves state-of-the-art performance on four public benchmarks of DML in both standard and \textit{Metric Learning Reality Check} (MLRC) \cite{musgrave2020metric} evaluation settings.
%\end{itemize}

\section{Related Work}

\noindent \textbf{Pair Mining / Pair Weighting in DML} \hspace{2mm}
% \subsection{Pair Mining / Pair Weighting in DML}
Non-uniform sampling \cite{DBLP:conf/icml/BengioLCW09,jiang2018mentornet,DBLP:journals/corr/AlainLSCB15,pmlr-v80-katharopoulos18a} and weighting \cite{DBLP:conf/nips/ChangLM17,8237586,NIPS2010_e57c6b95} methods have been shown to improve the performance of deep neural networks and have been widely applied to various tasks.
% As pair-based methods often generate redundant pairs, training with a uniform sampling can be overwhelmed by these \textit{easy} pairs, which significantly degrade the quality of the embedding space and also slows down the convergence.
% Therefore, 
In pair-based DML, many sampling strategies \cite{8237571,Duan_2019_CVPR} and a general pair weighting (GPW) framework \cite{wang2019multi} have been developed.

Hard sample mining is the most widely discussed sampling strategy for pair-based DML.
For example, Wu \etal \cite{8237571} showed that pair-wise distance distribution in the embedding space is biased, and proposed \textit{distance weighted sampling} to sample negative pairs uniformly according to the pair-wise distance within the mini-batch.
Additionally, the deep sampler network (DSN) \cite{Duan_2019_CVPR} was proposed to learn an adaptive sampling distribution based on the prior relations between training samples in the feature embedding network.
However, the architecture of DSN relies on the design of the loss function, and online sampling increases the training cost.
Wang \etal \cite{wang2019multi} proposed the general pair weighting (GPW) framework, which casts the sampling problem of deep metric learning into a unified view of pair weighting through gradient analysis.
They proposed the Multi-Similarity (MS) loss to generate non-uniform pair weights based on various pair relations.

Since hard mining is easily influenced by outliers that lead the learned model to a bad local optima \cite{Schroff_2015_CVPR}, methods such as Density Aware DML \cite{Ghosh_2019_CVPR} and Class-Aware Attention \cite{Wang_2019_AAAI} were proposed to identify outliers and reduce their impact on learning.
In Hardness-Aware DML \cite{Zheng_2019_CVPR}, the hardness threshold was scheduled by a global loss to perform adaptive mining.
Although hardness threshold and weighting score have been applied in existing methods, the proposed thresholds and weights were not adaptive to class hardness, which can vary during training.
Additionally, these methods provide pair-wise information by exploiting data-to-data relations within mini-batches, which results in the deficiency of global information; thus, the corresponding learning signal is sub-optimal.

\vspace{1mm}
\noindent \textbf{Proxy-Based Deep Metric Learning} \hspace{2mm}
% \subsection{Proxy-Based Deep Metric Learning}
% Most proxy-based methods introduce the concept of a proxy, which works as the representative of a class in the embedding space.
The use of proxy is raised by Proxy-NCA \cite{8237309}, which combines the NCA loss \cite{NIPS2004_42fe8808} with a proxy.
In its standard setting, one proxy is assigned to each class, and the raw data point is encouraged to be close to a positive proxy and far from all negative proxies.
Alternatively, Proxy-Anchor \cite{Kim_2020_CVPR} regards each proxy as an anchor, and intra-class relations are treated similarly to the MS loss \cite{wang2019multi}.
More recent studies \cite{qian2019striple,Zhu2020ProxyGML} also consider assigning more proxies to each class to capture intra-class variance.
For example, Qian \etal \cite{qian2019striple} extended the SoftMax loss to DML with multiple class centers, and proposed SoftTriple loss, which reflects the local geometry for each class.
ProxyGML \cite{Zhu2020ProxyGML} applies the proxy in a graph structure manner, and multiple proxies are selected to construct a subgraph to help raw data points learn the neighbor structure.
Although memory-based learning \cite{ko2021learning} was introduced to proxy-based methods for better generalization, existing methods do not consider the learning stability of the proxy itself, and still do not simultaneously address both intra- and inter-class relations in the loss function.

Rather than regarding the proxies as class representatives or graph nodes, the proposed Proxy-ISA treats each proxy as the center of a class-related region (subspace), and the definition of all the regions are estimated separately according to the proxies.
As the learning difficulty varies from class to class, the definition of regions also varies even within the same iteration.
In the proposed method, the proxy-data pairs are treated non-uniformly and adaptively based on the learned intra- and inter-class relations, resulting in better handling of global information.

% \vspace{1mm}
% \noindent \textbf{Active Learning} \hspace{2mm}
% % \subsection{Active Learning}
% In machine learning, active learning is a labeling strategy for datasets containing abundant unlabeled data, where informative unlabeled samples are prioritized.
% Previous studies \cite{settles2009active,DBLP:conf/nips/ChangLM17} have suggested that such a strategy is related to selecting informative samples during ordinary supervised learning.
% Active bias \cite{DBLP:conf/nips/ChangLM17} shows that neural networks can achieve better generalization by emphasizing samples with high prediction variance.
% Intuitively, if some samples have been already learned correctly with high confidence, \ie, close enough to its proxy, the pairs constructed from those samples may be too easy to contain useful information for further learning.
% However, if some samples are always predicted incorrectly over many iterations of training, \ie, too far from its proxy, they may be temporarily too difficult or noisy and may degrade the quality of the embedding space.
% This suggests that we should somehow prefer ``\textit{uncertain}" samples that stay within a certain interval of semantic distance from the proxy.%; Proxy-ISA applies two thresholds to detect such desired samples.

\section{Proposed Approach}

In this section, we first revisit the GPW framework \cite{wang2019multi} to provide a unified view of DML loss functions, and qualitatively discuss the impact and problems of hard samples on the learning of the proxies.
We then propose our gradient weighting method for proxy-based loss by introducing an estimation logic for hardness that adaptively changes along learning to ameliorate the problems.

\subsection{The Gradient Weighting Mechanism}

Let $\boldsymbol{x}_i \in \mathbb{R}^D$ be a real-value sample vector, and $y_i \in \{1, 2, \dots, C\}$ be the corresponding label.
Then the input matrix and the label vector for $m$ training samples can be denoted as $\boldsymbol{X} \in \mathbb{R}^{m \times D}$ and $\boldsymbol{y} \in \mathbb{R}^m$, respectively.
Let $\boldsymbol{p}_c \in \mathbb{R}^d$ be the proxy for class $c$ ($c = 1, 2, \dots, C$).
We denote a deep neural network by $f(\cdot; \boldsymbol{\theta}): \mathbb{R}^D \rightarrow \mathbb{R}^d$, where $\boldsymbol{\theta}$ denotes the corresponding parameters and $d$ is the embedding dimension.
Then the cosine similarity between a data embedding vector and a proxy can be defined as $\mathcal{S}_{i, c} := \langle f(\boldsymbol{x}_i; \boldsymbol{\theta}), \boldsymbol{p}_c \rangle$, where $\langle\cdot, \cdot\rangle$ denotes the dot product.
Similarly, the similarity matrix can be denoted as $\boldsymbol{S} \in \mathbb{R}^{m \times C}$.

Given a proxy-based loss $\mathcal{L}(\boldsymbol{S}, \boldsymbol{y})$, the derivative \wrt\ $\boldsymbol{\theta}$ at the $t$-th iteration can be written as
\begin{equation}
    \begin{split}
        \frac{\partial\mathcal{L}(\boldsymbol{S}, \boldsymbol{y})}{\partial\boldsymbol{\theta}}\bigg|_t &= \sum_{i = 1}^m \sum_{c = 1}^C \frac{\partial\mathcal{L}(\boldsymbol{S}, \boldsymbol{y})}{\partial \mathcal{S}_{i, c}}\bigg|_t \frac{\partial \mathcal{S}_{i, c}}{\partial\boldsymbol{\theta}}\bigg|_t\\
        &= \sum_{i = 1}^m \left( \sum_{c \neq y_i} \frac{\partial\mathcal{L}}{\partial \mathcal{S}_{i, c}}\bigg|_t \frac{\partial \mathcal{S}_{i, c}}{\partial\boldsymbol{\theta}}\bigg|_t + \sum_{c = y_i} \frac{\partial\mathcal{L}}{\partial \mathcal{S}_{i, c}}\bigg|_t \frac{\partial \mathcal{S}_{i, c}}{\partial\boldsymbol{\theta}}\bigg|_t \right).
    \end{split}
    \label{eq1}
\end{equation}
$\frac{\partial\mathcal{L}}{\partial \mathcal{S}_{i, c}}\big|_t$ in \cref{eq1} is regarded as a constant scalar in the gradient \wrt\ $\boldsymbol{\theta}$.
Since positive pairs are encouraged to be close and negative pairs need to be pushed away from each other, we assume $\frac{\partial\mathcal{L}}{\partial \mathcal{S}_{i, c}}\big|_t \geq 0$ for negative pairs, and $\frac{\partial\mathcal{L}}{\partial \mathcal{S}_{i, c}}\big|_t \leq 0$ for positive pairs.
Thus, \cref{eq1} can be transformed into a form of weighted sum:
\begin{equation}
    \frac{\partial\mathcal{L}(\boldsymbol{S}, \boldsymbol{y})}{\partial\boldsymbol{\theta}}\bigg|_t = \sum_{i = 1}^m \left( \sum_{c \neq y_i} w_{i, c} \frac{\partial \mathcal{S}_{i, c}}{\partial\boldsymbol{\theta}}\bigg|_t - \sum_{c = y_i} w_{i, c} \frac{\partial \mathcal{S}_{i, c}}{\partial\boldsymbol{\theta}}\bigg|_t \right),
    \label{eq2}
\end{equation}
where $w_{i, c} = \left|\frac{\partial\mathcal{L}(\boldsymbol{S}, \boldsymbol{y})}{\partial \mathcal{S}_{i, c}}\right|_t$.
\cref{eq2} suggests that the gradient signal is actually controlled by $w_{i, c}$, namely, how the similarity metric is defined in the DML loss function.

\subsection{Impact of Gradient Weights on Proxies}

Let $\boldsymbol{P} \in \mathbb{R}^{C \times d}$ be the proxy set.
% If we focus on the learning of a proxy, \ie\ the gradient \wrt\ $\boldsymbol{P}$, we can rewrite \cref{eq2} by replacing $\boldsymbol{\theta}$:
The gradient \wrt\ $\boldsymbol{P}$ can be obtained by replacing $\boldsymbol{\theta}$ in \cref{eq2}:
\begin{equation}
    \frac{\partial\mathcal{L}(\boldsymbol{S}, \boldsymbol{y})}{\partial\boldsymbol{P}}\bigg|_t = \sum_{i = 1}^m \left( \sum_{c \neq y_i} w_{i, c} \frac{\partial \mathcal{S}_{i, c}}{\partial\boldsymbol{p}_c}\bigg|_t - \sum_{c = y_i} w_{i, c} \frac{\partial \mathcal{S}_{i, c}}{\partial\boldsymbol{p}_c}\bigg|_t \right).
    \label{eq3}
\end{equation}
\cref{eq3} suggests that the learning of a proxy is also affected by $w_{i, c}$.
In pair-based methods, the effect of $w_{i, c}$ is symmetrical since we can treat either side of the pair as an anchor.
However, in proxy-based methods, larger gradient weights may result in an undesirable proxy distribution, which is detrimental to the representation of global information.

The first proxy-based loss combined with gradient weighting was the Proxy-Anchor loss \cite{Kim_2020_CVPR}, which is formulated as follows:
% \scalebox{0.8}{
\begin{equation}
\begin{split}
    \mathcal{L}_{PA} :=& \frac{1}{|\boldsymbol{C}^+|}\sum_{c \in \boldsymbol{C}^+} \log\left( 1 + \sum_{i: y_i = c} e^{\alpha(\delta - \mathcal{S}_{i, c})} \right)\\
                       & + \frac{1}{C}\sum_{c = 1}^C \log\left( 1 + \sum_{i: y_i \neq c} e^{\alpha(\delta + \mathcal{S}_{i, c})} \right),
\end{split}
    \label{proxy_anchor}
\end{equation}
% }
where $\boldsymbol{C}^+ = \{c | c \in \boldsymbol{y}\}$ denotes a set of class labels in a mini-batch, and $\alpha$, $\delta$ are fixed hyper-parameters.
By taking the derivative of $\mathcal{L}_{PA}$ \wrt\ $\mathcal{S}_{i, c}$, the gradient weights are given by
\begin{equation}
    w_{i, c} = \begin{cases}
        \frac{1}{|\boldsymbol{C}^+|}\frac{\alpha e^{\alpha(\delta - \mathcal{S}_{i, c})}}{1 + \underset{j: y_j = c}{\sum} e^{\alpha(\delta - \mathcal{S}_{j, c})}}, & y_i = c,\\
        \frac{1}{C}\frac{\alpha e^{\alpha(\delta + \mathcal{S}_{i, c})}}{1 + \underset{j: y_j \neq c}{\sum} e^{\alpha(\delta + \mathcal{S}_{j, c})}}, & y_i \neq c.
    \end{cases}
    \label{weight_pa}
\end{equation}
From \cref{weight_pa}, it is clear that for a positive pair, $w_{i, c}$ will be large if the data embedding is far from its proxy (\ie\ $\mathcal{S}_{i, c}$ is small), and will be larger if it is farther than other positive pairs related to the same proxy (\ie\ $\mathcal{S}_{i, c} < \forall \mathcal{S}_{j, c}$).
A similar approach is adopted for the negative pairs, as illustrated in \cref{fig1.1}, where the thickness of the arrows indicates the magnitude of the gradient weights.
If a positive pair contains an outlier, the proxy will be pulled heavily in that direction.

We now analyze the optimization problem according to each term in the Proxy-Anchor loss.

\begin{figure}[t]
    \centering
    \begin{subfigure}[b]{0.49\linewidth}
        \centering
        \includegraphics[width=\linewidth]{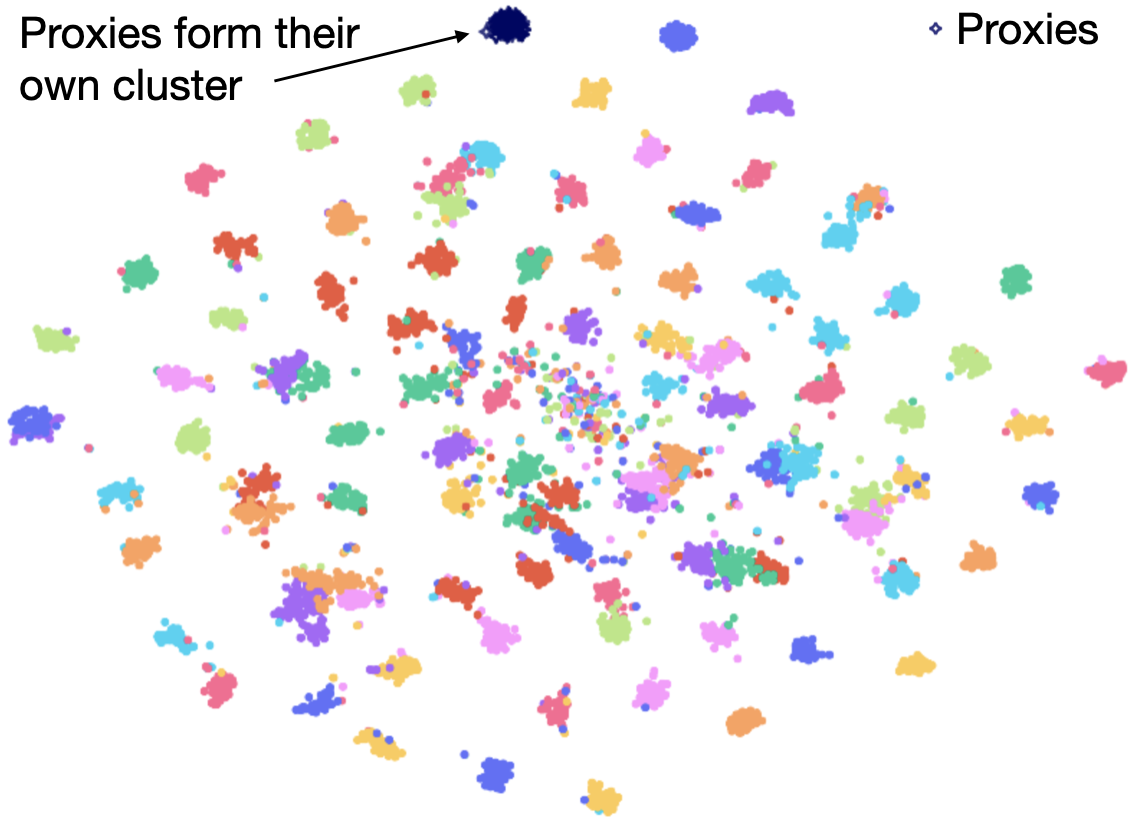}
        \caption{Proxy-Anchor@epoch30}
        \label{fig3.1a}
    \end{subfigure}
    \hfill
    \begin{subfigure}[b]{0.49\linewidth}
        \centering
        \includegraphics[width=\linewidth]{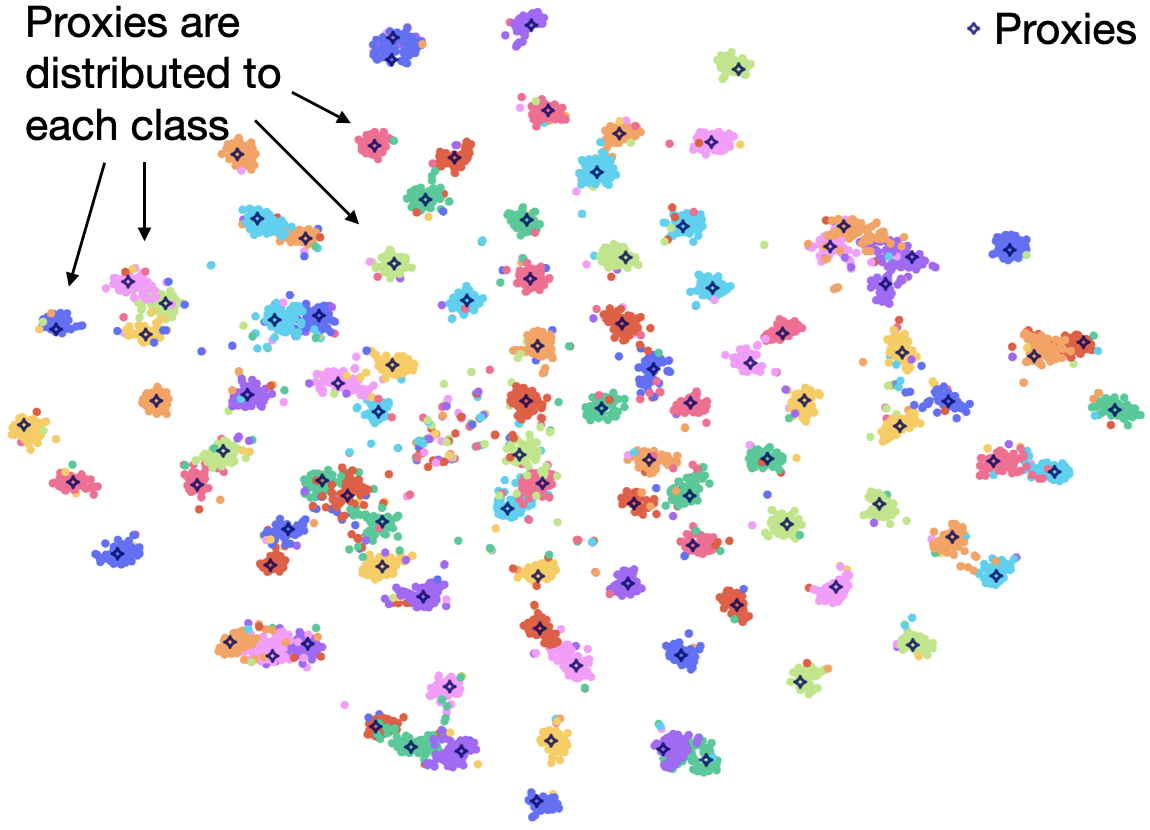}
        \caption{Norm.SoftMax@epoch30}
        \label{fig3.1b}
    \end{subfigure}
    \caption{t-SNE \cite{vanDerMaaten2008} visualization of 512-dimensional embedding space for the Cars-196 dataset. (a) and (b) present the embedding space learned by Proxy-Anchor and normalized SoftMax, respectively. The proxies of Proxy-Anchor, plotted with navy blue stars, form a cluster at an early stage due to the strong repulsive forces.}
    \label{fig3.1}
\end{figure}

% \subsubsection{Proposition 1.}
\vspace{1mm}
\noindent \textbf{Proposition 1} \hspace{2mm}
\begin{equation}
\begin{split}
    \log\left( 1 + \sum_{i: y_i = c} e^{\alpha(\delta - \mathcal{S}_{i, c})} \right) &= \max_{\mathcal{P}_c^+} \alpha \sum_{i: y_i = c} \mathcal{P}_c^+(i)(\delta - \mathcal{S}_{i, c}) + H(\mathcal{P}_c^+),\\
    \log\left( 1 + \sum_{i: y_i \neq c} e^{\alpha(\delta + \mathcal{S}_{i, c})} \right) &= \max_{\mathcal{P}_c^-} \alpha \sum_{i: y_i \neq c} \mathcal{P}_c^-(i)(\mathcal{S}_{i, c} + \delta) + H(\mathcal{P}_c^-),
\end{split}
    \label{prop1}
\end{equation}
\textit{where $H(\cdot)$ denotes entropy for regularization, $\mathcal{P}_c^+$ and $\mathcal{P}_c^-$ are the probability distributions over the positives and negatives related to a proxy $\boldsymbol{p}_c$, respectively, and $\mathcal{P}_c^+(i) \in \{\mathcal{P}_c^+ | \mathcal{P}_c^+(\delta) + \sum_{i: y_i = c} \mathcal{P}_c^+(i) = 1, \forall i, \mathcal{P}_c^+(i) \geq 0\}$, $\mathcal{P}_c^-(i) \in \{\mathcal{P}_c^- | \mathcal{P}_c^-(-\delta) + \sum_{i: y_i \neq c} \mathcal{P}_c^-(i) = 1, \forall i, \mathcal{P}_c^-(i) \geq 0\}$, where we define}
\begin{equation}
    \mathcal{P}_c^+(\delta) = \frac{1}{1 + \underset{j: y_j = c}{\sum} e^{\alpha(\delta - \mathcal{S}_{j, c})}}, \quad
    \mathcal{P}_c^-(-\delta) = \frac{1}{1 + \underset{j: y_j \neq c}{\sum} e^{\alpha(\mathcal{S}_{j, c} - (-\delta))}}
\label{def_pDelta}
\end{equation}
% \textit{Proof.} According to the K.K.T. conditions \cite{boyd2004convex}, $\mathcal{P}_c^+$ in \cref{prop1} has the closed-form solution
% \begin{equation}
%     \mathcal{P}_c^+(i) = \frac{e^{\alpha(\delta - \mathcal{S}_{i, c})}}{1 + \underset{j: y_j = c}{\sum} e^{\alpha(\delta - \mathcal{S}_{j, c})}}.
% \end{equation}
% Therefore, we have
% \begin{equation}
%     \alpha \sum_{i: y_i = c} \mathcal{P}_c^+(i)(\delta - \mathcal{S}_{i, c}) + H(\mathcal{P}_c^+) = \log\left( 1 + \sum_{i: y_i = c} e^{\alpha(\delta - \mathcal{S}_{i, c})} \right).
% \end{equation}
% The same analysis is applicable to $\mathcal{P}_c^-$.
(see supplementary material for proof).

Proposition 1 suggests that Proxy-Anchor loss actually optimizes two data distributions related to each proxy, and that these distributions are independent to each other.
Thus, the global proxy distribution is not relevant during training.
This is in contrast to the normalized SoftMax loss, which considers an optimal global proxy distribution related to each data embedding \cite{qian2019striple}, resulting in updates of proxies that are completely dominated by data embeddings.
Since more samples are regarded as hard in the early stages of learning, the repulsive force generated by negatives cause the proxies to be far from the ideal distribution, as illustrated in \cref{fig3.1}.
Additionally, solutions for the positive terms are sub-optimal when the positives contain outliers.

\begin{figure}[t]
    \centering
    \includegraphics[width=\linewidth]{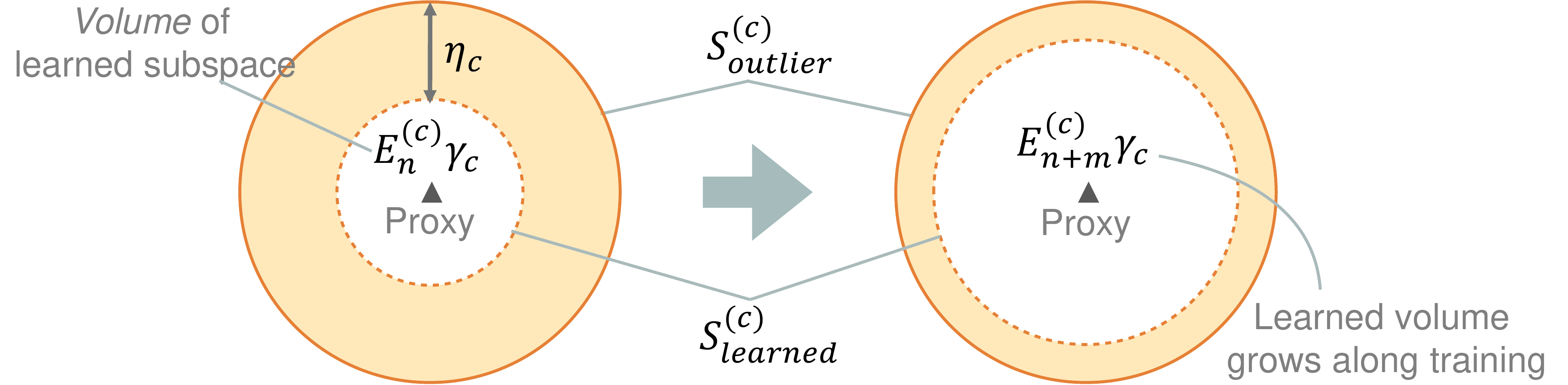}
    \caption{Two thresholds related to a proxy. $\boldsymbol{\mathcal{S}_{outlier}}$: similarity threshold to detect outliers, \ie\ data embeddings with $\mathcal{S}_{i, c} < \mathcal{S}_{outlier}^{(c)}$; $\boldsymbol{\mathcal{S}_{learned}}$: similarity threshold to detect overlapping between the learned subspace and the newly sampled data. Samples that satisfy $\mathcal{S}_{i, c} > \mathcal{S}_{learned}^{(c)}$ are considered too easy for the model. When $E_n^{(c)}$ increases, $\eta_c$ should decrease.}
    \label{fig3.3}
\end{figure}

\subsection{Class Hardness}
\label{sec3.3}

Following the idea of active learning, we can reduce the weights of outliers to stabilize the learning of the proxy if most of the data embeddings are close enough to the proxies (\ie\ easy classes).
The problem is then how to obtain a boundary to detect outliers for each class.
% If a class is regarded as difficult, the corresponding outlier boundary should be made wider to reserve a wider subspace for better discriminative capability.
We define such an outlier boundary as a class-(proxy-) dependent similarity threshold $\mathcal{S}_{outlier}^{(c)}$, and define $\eta_c$ to be the range for informative samples.
In other words, we increase the gradient weights for positive pairs such that $\mathcal{S}_{i, c} \in [\mathcal{S}_{outlier}^{(c)}, \mathcal{S}_{outlier}^{(c)} + \eta_c]$ and decrease them otherwise, where $\mathcal{S}_{outlier}^{(c)} + \eta_c$ should be another threshold related to the current learned subspace for class $c$.

Since we do not have any \textit{a priori} knowledge about the difficulty of each class, all samples should be treated equally in the early stages of training.
Such distinctions should be made independently after the subspace of the class has reached a certain learning stage.

\vspace{1mm}
\noindent \textbf{Volume of the learned subspace} \hspace{2mm}
To simulate the learning of the subspace, we assume each data embedding owns one \textit{volumetric unit} of the subspace belonging to its class, and apply the theory of \textit{effective number} (discussed in \cite{8953804}).

Let $\gamma_c$ be the volumetric unit for class $c$, we assume it is affected by the class hardness and should be larger for a difficult class.
A sample from a harder class contains more information than that from an easier class.
This is consistent with the fact that most data embeddings of easier classes are closer to each other than that of harder classes.
According to the original definition in \cite{8953804}, the total volume of the feature space for each class has an upper bound, and random sampling can be considered as randomly covering this volume.
This means that overlapping between the samples randomly happens.
Let $V\gamma_c$ be the upper bound of the total volume and $\beta = \frac{V - 1}{V}$.
Then the effective number (denoted as $E_n$) is defined as
\begin{equation}
    E_n^{(c)}\gamma_c = \frac{1 - \beta^n}{1 - \beta} \gamma_c,
    \label{defE_n}
\end{equation}
where
\begin{equation}
    \lim_{n \rightarrow \infty}E_n^{(c)} = \frac{1}{1 - \beta} = V
    \label{limE_n}
\end{equation}
holds.
\cref{defE_n,limE_n} suggest that $E_n$ is independent from the class hardness.
Therefore, in the proposed approach, we regard the effective number as training \textit{progress bar} and introduce an effective number for each class.
The upper bound $V$ is set as a hyper-parameter.
Furthermore, we treat the boundary of the feature space as the outlier boundary, and define $\mathcal{S}_{outlier}^{(c)} + \eta_c$ as the threshold of the learned subspace related to class $c$, denoted as $\mathcal{S}_{learned}^{(c)}$.
Hence, $\eta_c$ decreases as the learning progresses, as shown in \cref{fig3.3}.

\vspace{1mm}
\noindent
\textbf{Estimating the Class Hardness} \hspace{2mm}
% Let $\gamma_c$ be the volumetric unit for class $c$, then the volume of the learned set of this class with size $n$ can be expressed as $E_n^{(c)}\gamma_c$.
% From the above definition, $\gamma_c$ is affected by the class hardness; $\gamma_c$ should be larger for a difficult class.
% A sample from a harder class contains more information than that from an easier class.
% This is consistent with the fact that most data embeddings of easier classes are closer to each other than that of harder classes.
Based on the above definition, for a proxy-based method, the average cosine similarity of the learned positive proxy-data pairs (excluding the outliers) reflects how well the model learned about a class, and can be used to estimate the class hardness after a certain stage.

\begin{figure}[t]
    \centering
    \includegraphics[width=\linewidth]{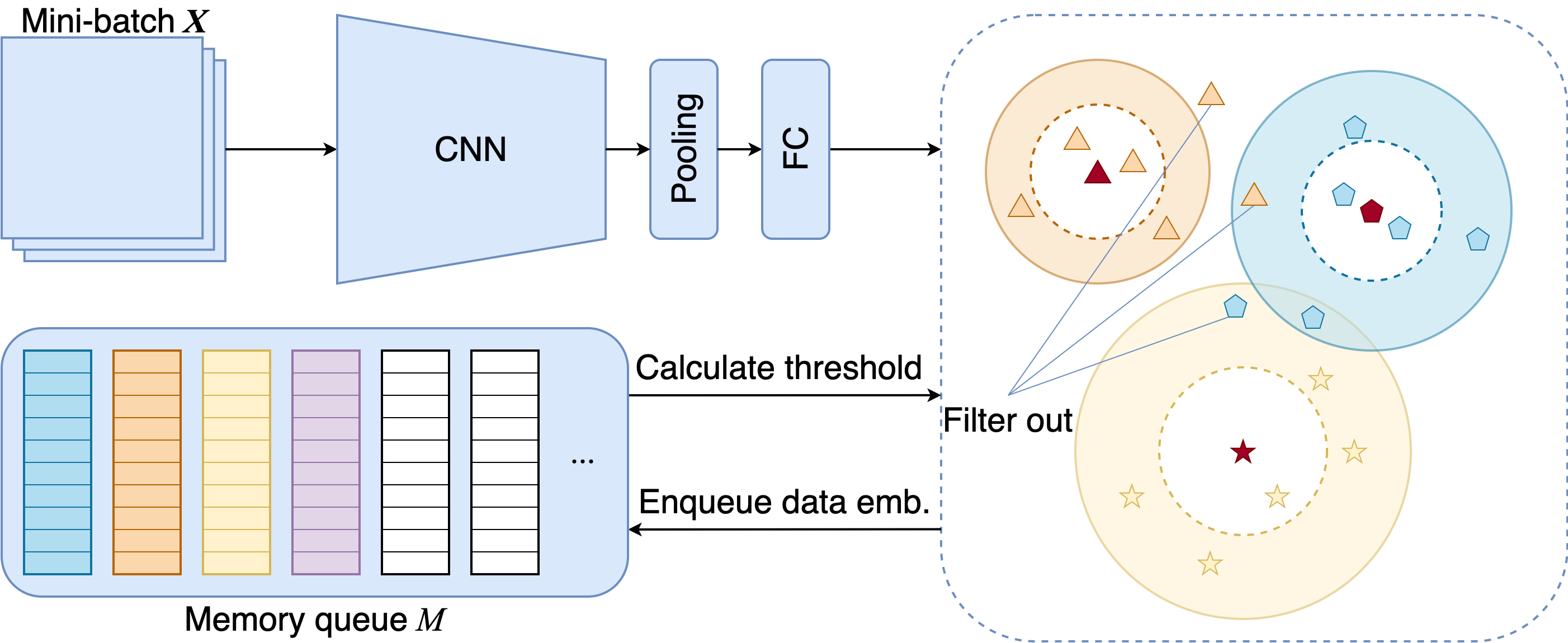}
    \caption{Training flow of Proxy-ISA. Shapes denote embeddings from different classes, whose thresholds are denoted in different colors. When the size of $M$ reaches $T$, it dequeues the old embeddings to keep the memory up-to-date. Embeddings from the past few iterations are available for reference due to the \textit{slow drift} phenomenon discussed in \cite{wang2020xbm}. The algorithm is described in supplementary material.}
    \label{fig3.4}
\end{figure}

Estimating the class hardness using only the current mini-batch may not be enough, since there are usually only a few samples from the same class in the mini-batch, and the number of samples for some classes may be zero.
Therefore, we apply a memory queue to store the clean data embeddings which were sampled during the past few iterations.
Formally, let $M$ be the memory queue with max size $T$, and $T_c$ be the total number of embeddings from class $c$ in $M$.
Then $\mathcal{S}_{learned}^{(c)}$ is calculated by
\begin{equation}
    \mathcal{S}_{learned}^{(c)} = h \cdot \mathcal{S}_{avg}^{(c)} = \frac{h}{T_c}\sum_{i: (\boldsymbol{x}_i, y_i) \in M, y_i = c} \mathcal{S}_{i, c},
    \label{defOverlapSim}
\end{equation}
where $h$ is the hyper-parameter used to scale the hardness.
After processing a mini-batch, we filter out outliers and enqueue the rest to $M$, as illustrated in \cref{fig3.4}.
The memory queue starts after a certain number of steps is reached, because the embeddings in the initial stages are typically scattered and the resulting similarities are not representative.

\subsection{Informative Sample-Aware Proxy}

Our proposed Proxy-ISA utilizes the aforementioned properties to generate dynamic pair-weights.
Let $\omega_{i, c}$ be the weighting score for the pair which consists of a data embedding $f(\boldsymbol{x}_i; \boldsymbol{\theta})$ and a proxy $\boldsymbol{p}_c$.
This weighting score depends on the semantic state of $\boldsymbol{x}_i$ and learning progress $E_n^{(c)}$.
Let $\omega_{i, c}^+$ and $\omega_{i, c}^-$ denote the weighting score for positive pairs and negative pairs, respectively.

To reduce the harmful factors from extremely hard samples, we first reform the optimization problems as
\begin{equation}
\begin{split}
    \max_{\mathcal{Q}_c^+} \alpha \sum_{i: y_i = c} \omega_{i, c}^+\mathcal{Q}_c^+(i)(\delta - \mathcal{S}_{i, c}) + H(\mathcal{Q}_c^+)\\
    \max_{\mathcal{Q}_c^-} \alpha \sum_{i: y_i \neq c} \omega_{i, c}^-\mathcal{Q}_c^-(i)(\mathcal{S}_{i, c} + \delta) + H(\mathcal{Q}_c^-)
\end{split}
\label{eqdefQ}
\end{equation}
where $\mathcal{Q}_c^+$ and $\mathcal{Q}_c^-$ have the same form as $\mathcal{P}_c^+$ and $\mathcal{P}_c^-$, respectively.
According to Proposition 1, the optimal of the formulas in \cref{eqdefQ} are
\begin{equation}
    \log\left( 1 + \sum_{i: y_i = c} e^{\omega_{i, c}^+\cdot\alpha(\delta - \mathcal{S}_{i, c})} \right), \quad
    \log\left( 1 + \sum_{i: y_i \neq c} e^{\omega_{i, c}^-\cdot\alpha(\delta + \mathcal{S}_{i, c})} \right),
\end{equation}
respectively.
Thus, the objective function of Proxy-ISA can be formulated as
\begin{equation}
    \begin{split}
        \mathcal{L}_{ISA} :=& \frac{1}{\underset{c \in \boldsymbol{C}^+}{\sum}\bar{\omega}_c^+} \sum_{c \in \boldsymbol{C}^+}\log\left( 1 + \sum_{i: y_i = c} e^{\omega_{i, c}^+\cdot\alpha(\delta - \mathcal{S}_{i, c})} \right)\\
        &+ \frac{1}{\underset{c}{\sum}\bar{\omega}_c^-} \sum_{c = 1}^C \log\left( 1 + \sum_{i: y_i \neq c} e^{\omega_{i, c}^-\cdot\alpha(\delta + \mathcal{S}_{i, c})} \right),
    \end{split}
    \label{loss_pisa}
\end{equation}
where $\bar{\omega}_c^+$ and $\bar{\omega}_c^-$ are the average of all $\omega_{i, c}^+$ and $\omega_{i, c}^-$, respectively, for all samples $i$ that belong to class $c$.
The summation reflects the average scores of all classes (the global learning status) that relate to the positive or negative term.
% The corresponding gradient weights are then given by
% \begin{equation}
%     w_{i, c}^* = \begin{cases}
%         \frac{\omega_{i, c}^+}{\underset{c^\prime \in \boldsymbol{C}^+}{\sum}\bar{\omega}_{c^\prime}^+}\frac{\alpha e^{-\omega_{i, c}^+\cdot\alpha(\mathcal{S}_{i, c} - \delta)}}{1 + \underset{j: y_j = c}{\sum} e^{-\omega_{jc}^+\cdot\alpha(\mathcal{S}_{j, c} - \delta)}}, & y_i = c,\\
%         \frac{\omega_{i, c}^-}{\underset{c^\prime}{\sum}\bar{\omega}_{c^\prime}^-}\frac{\alpha e^{\omega_{i, c}^-\cdot\alpha(\mathcal{S}_{i, c} + \delta)}}{1 + \underset{j: y_j \neq c}{\sum} e^{\omega_{jc}^-\cdot\alpha(\mathcal{S}_{j, c} + \delta)}}, & y_i \neq c.
%     \end{cases}
%     \label{weight_pisa}
% \end{equation}

% \subsubsection{Adaptive Pair Weighting Scores}
\vspace{1mm}
\noindent \textbf{Adaptive Pair Weighting Scores} \hspace{2mm}
Since the model has higher confidence in the intra-class hardness when the learning progresses to a later stage, the penalty from $\omega_{i, c}$ is greater for a larger $E_n^{(c)}$.
A naive implementation is to set $\omega_{i, c} = \frac{1}{E_n^{(c)}}$ for both positive and negative pairs when penalty is needed.
However, this causes the learning signal to be too weak for positive pairs and thus an imbalance between the positive and negative terms when $E_n^{(c)}$ is large, since the penalties for outliers are only made for positive pairs. %(\ie, more penalties compared to the negative pairs).
To prevent the diminishing of $\omega_{i, c}^+$, we set a lower bound $\nu_n^{(c)}$ for all positive pairs, defined as
\begin{equation}
    \nu_n^{(c)} = \frac{1}{1 + \log(1 + E_n^{(c)})},
    \label{defOmega_base}
\end{equation}
% where $\nu_0^{(c)} = 1$.
% From \cref{limE_n}, $\lim_{n \to \infty} \nu_n^{(c)} = \frac{1}{1 + \log(1 + V)}$ holds.

Based on the analysis in \cref{sec3.3}, we define $\eta_c$ as follows:
\begin{equation}
    \eta_c = (1 + k(1 - h \cdot \mathcal{S}_{avg}^{(c)}))\nu_n^{(c)} + \lambda,
    \label{defEta}
\end{equation}
where $k$ ($> 0$) is the sensitivity factor and $\lambda \in [0, 1]$ is the margin.
Thus, the searching length $\eta_c$ is controlled by class hardness, where a harder class has a wider searching space, and an easier class is restricted to help stabilize the learning.

\begin{figure}[t]
    \centering
    \includegraphics[width=\linewidth]{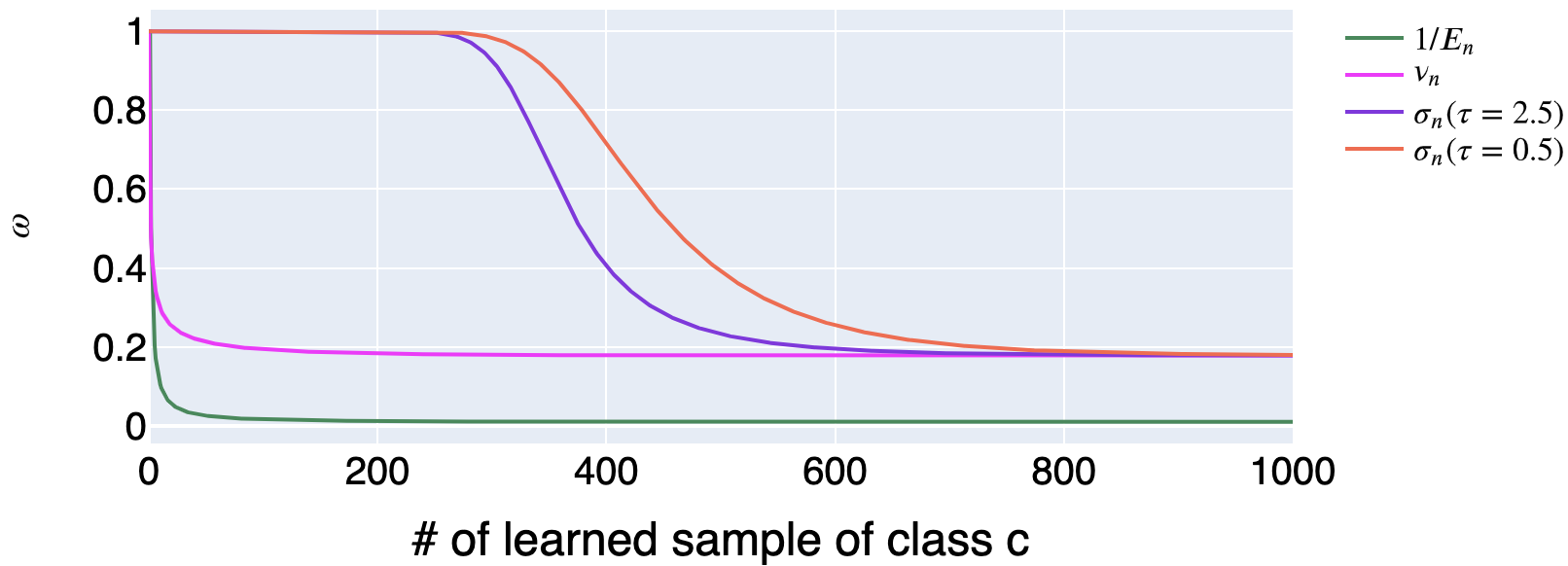}
    \caption{Curve of the decay functions.}
    \label{fig3.5}
    \vspace{-2mm}
\end{figure}

Finally, the proposed dynamic weights are defined as follows:
\begin{equation}
    \begin{split}
        \omega_{i, c}^+ &= \begin{cases}
            1 + \sigma_n^{(c)}, & h \cdot \mathcal{S}_{avg}^{(c)} - \eta_c \leq \mathcal{S}_{i, c} \leq h \cdot \mathcal{S}_{avg}^{(c)},\\
            \sigma_n^{(c)}, & \text{otherwise},
        \end{cases}\\
        \omega_{i, c}^- &= \begin{cases}
            \frac{1}{\max(1, E_n^{(c)})}, & \mathcal{S}_{i, c} < h \cdot \mathcal{S}_{avg}^{(c)} - \eta_c,\\
            1, & \text{otherwise},
        \end{cases}
    \end{split}
    \label{defOmega}
\end{equation}
where $\sigma_n^{(c)}$ is a decay function for $\omega_{i, c}^+$, defined as
% This suggests that we need a function with a concave curve in the initial phase, such as the sigmoid function, which forms basis of $\sigma_n^{(c)}$ (see supplementary material for derivation):
% Hence, $\sigma_n^{(c)}$ is designed based on the sigmoid function (see supplementary material for derivation):
\begin{equation}
    \sigma_n^{(c)} = 1 + \frac{(1 + e^{-\tau})(\nu_n^{(c)} - 1)}{1 + e^{V - E_n^{(c)} - \tau}},
    \label{defSigma}
\end{equation}
and $\tau$ ($> 0$) is a hyper-parameter that controls the timing of decay.
Since the embeddings are more likely to change in the early stages and gradually stabilize later, the dynamic weights need to be controlled such that they change less in the early stages (see supplementary material for derivation of $\sigma_n^{(c)}$).
Using \cref{limE_n} and \cref{defOmega_base}, $\lim_{n \to \infty}\sigma_n^{(c)} = \lim_{n \to \infty}\nu_n^{(c)} = \frac{1}{1 + \log(1 + V)}$.
This ensures that $\omega_{i, c}^+$ does not fall below $\nu_n^{(c)}$, as shown in \cref{fig3.5}.

\section{Experiments}

We conduct experiments on four widely used benchmarks to evaluate and analyze the effectiveness of Proxy-ISA.

\subsection{Experimental Setting}

% \subsubsection{Datasets}
\noindent
\textbf{Datasets} \hspace{2mm}
Experiments were conducted on the CUB-200-2011 \cite{WahCUB_200_2011}, Cars-196 \cite{6755945}, Stanford Online Products (SOP) \cite{song2016deep}, and In-Shop Clothes Retrieval (In-Shop) \cite{Liu_2016_CVPR} datasets.
We follow the standard protocol applied in \cite{song2016deep} to split them into training and testing parts.
CUB-200-2011 contains 200 species of birds with 11,788 images; we used the first 100 classes (5,864 images) for training and the rest for testing.
Cars-196 consists of 196 model categories of cars with 16,185 images; the first 98 classes (8,054 images) were used for training and the rest are used for testing.
SOP contains 22,634 classes with 120,053 product images; we used the first 11,318 classes (59,551 images) for training and the rest for testing.
The first 3,997 classes (25,882 images) of In-Shop were used for training, while the remaining 3,985 classes were used for the test set, which was partitioned into a query set and a gallery set containing 14,218 and 12,612 images, respectively.

\vspace{1mm}
\noindent
\textbf{Evaluation Metrics} \hspace{2mm}
The evaluation procedure included two types of metrics.
We first performed a standard evaluation following \cite{song2016deep}, calculating the Recall@$K$ for image retrieval tasks.
We also adopted MAP@R, known as \textit{Mean Average Precision}, from the MLRC \cite{musgrave2020metric} evaluation settings.
MAP@R considers the correctness of each result associated with a single query, and is considered to be more suitable for evaluating the entire embedding space.

\begin{figure}[t]
    \centering
    \begin{subfigure}[b]{0.49\linewidth}
        \centering
        \includegraphics[width=\linewidth]{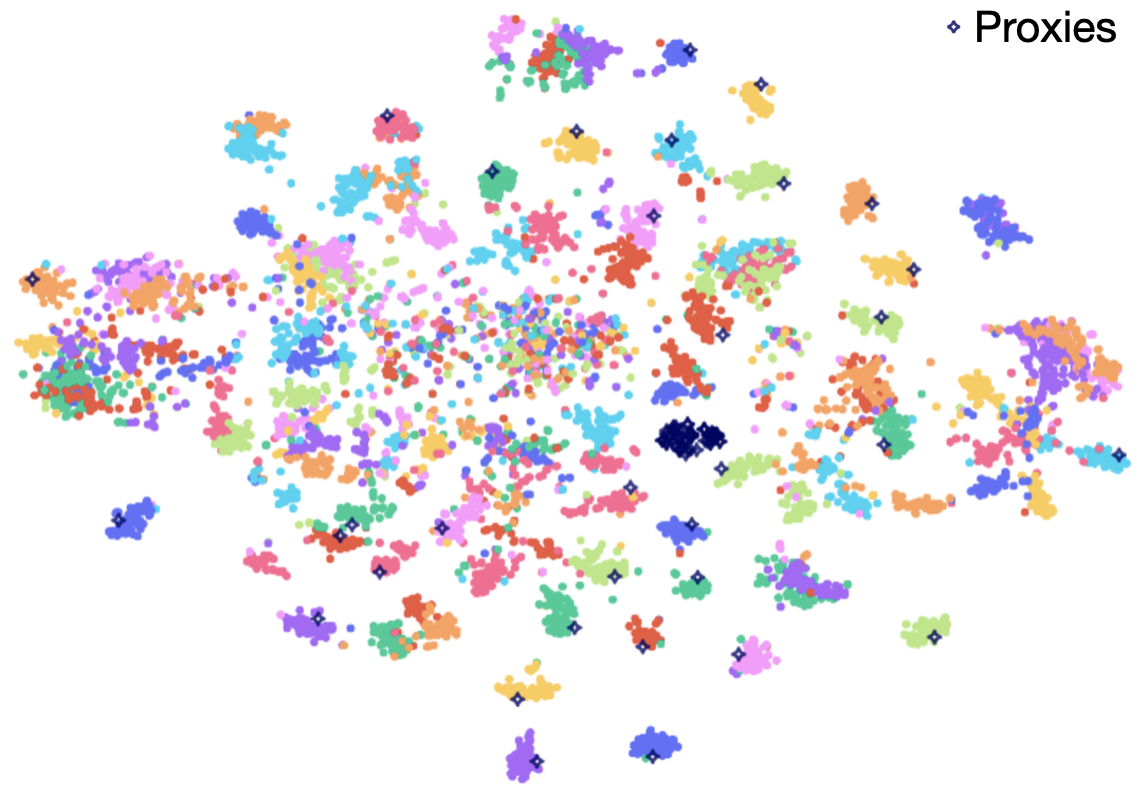}
        \caption{P-ISA@epoch30}
        \label{fig4.1a}
    \end{subfigure}
    \hfill
    \begin{subfigure}[b]{0.49\linewidth}
        \centering
        \includegraphics[width=\linewidth]{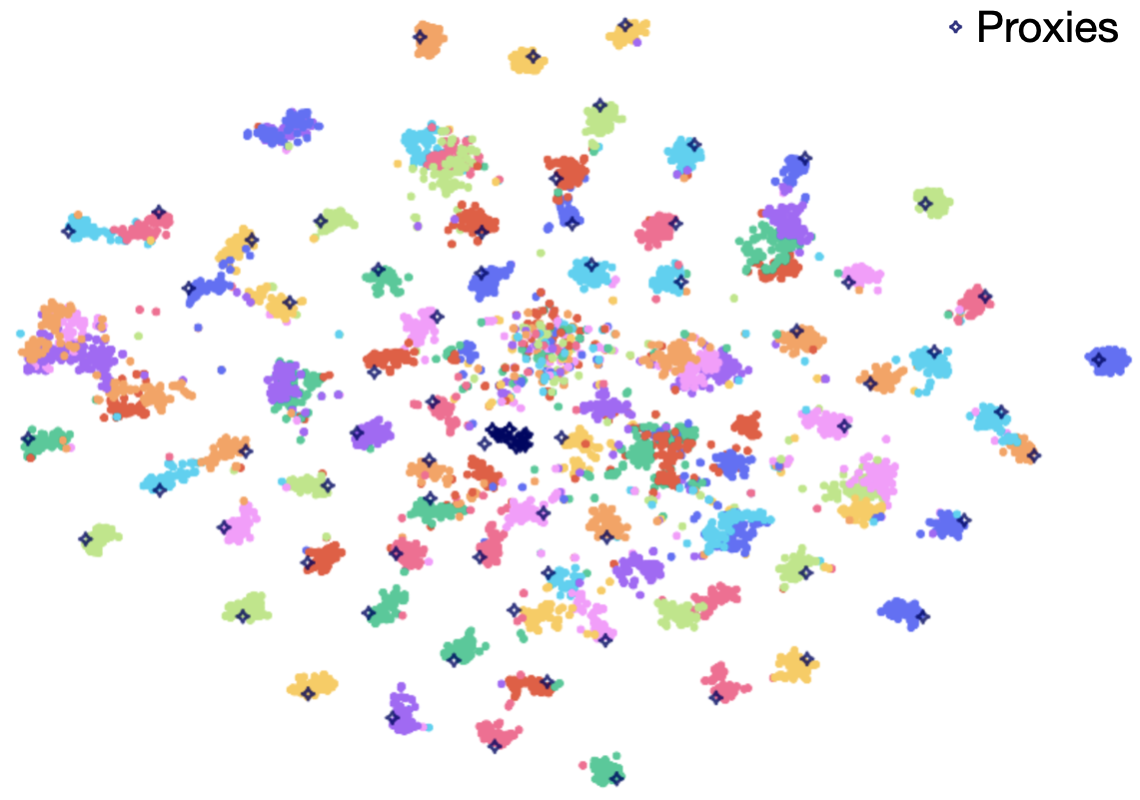}
        \caption{P-ISA@epoch80}
        \label{fig4.1b}
    \end{subfigure}
    % \hfill
    \begin{subfigure}[b]{0.49\linewidth}
        \centering
        \includegraphics[width=\linewidth]{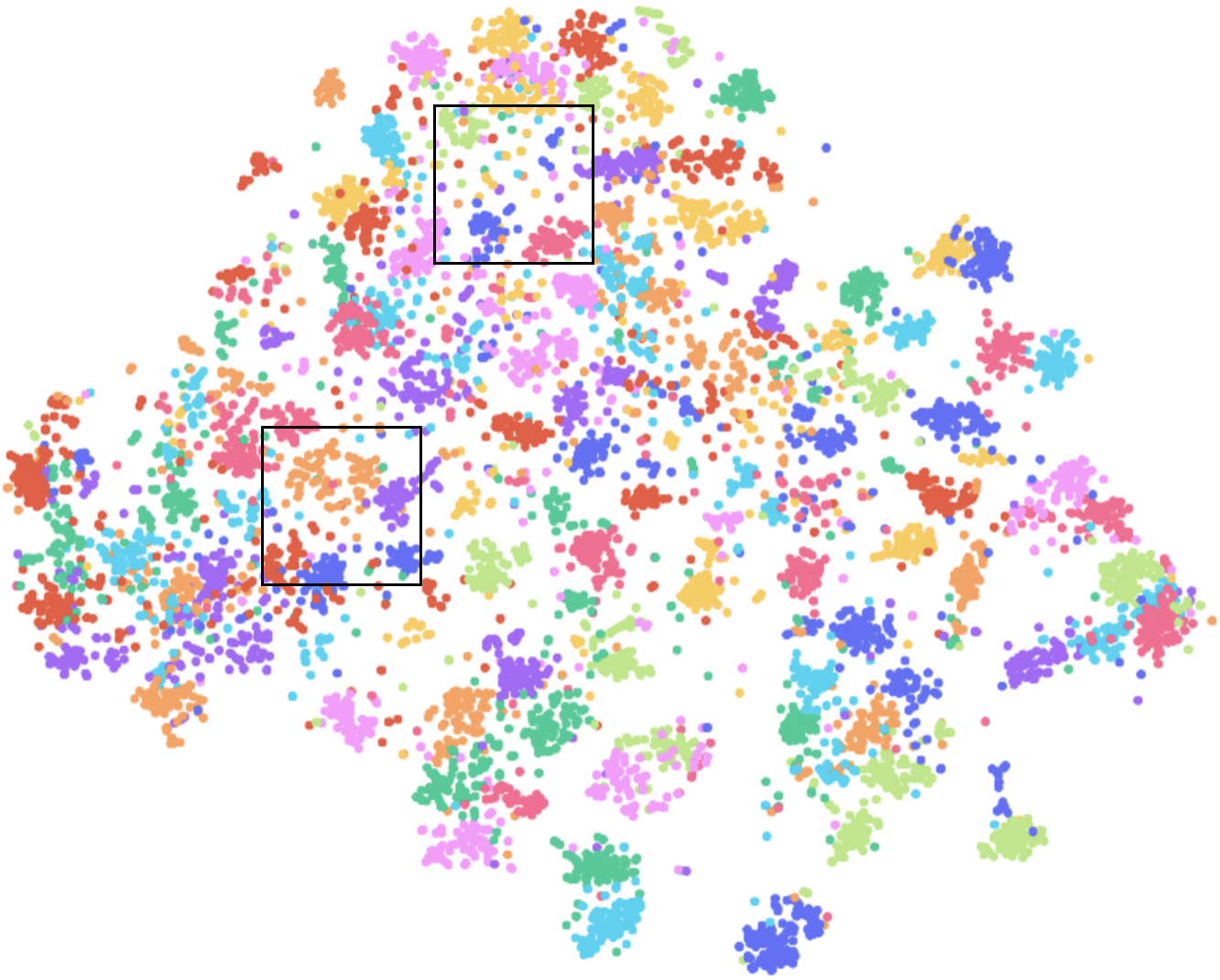}
        \caption{Test set, PA}
        \label{fig4.1c}
    \end{subfigure}
    \hfill
    \begin{subfigure}[b]{0.49\linewidth}
        \centering
        \includegraphics[width=\linewidth]{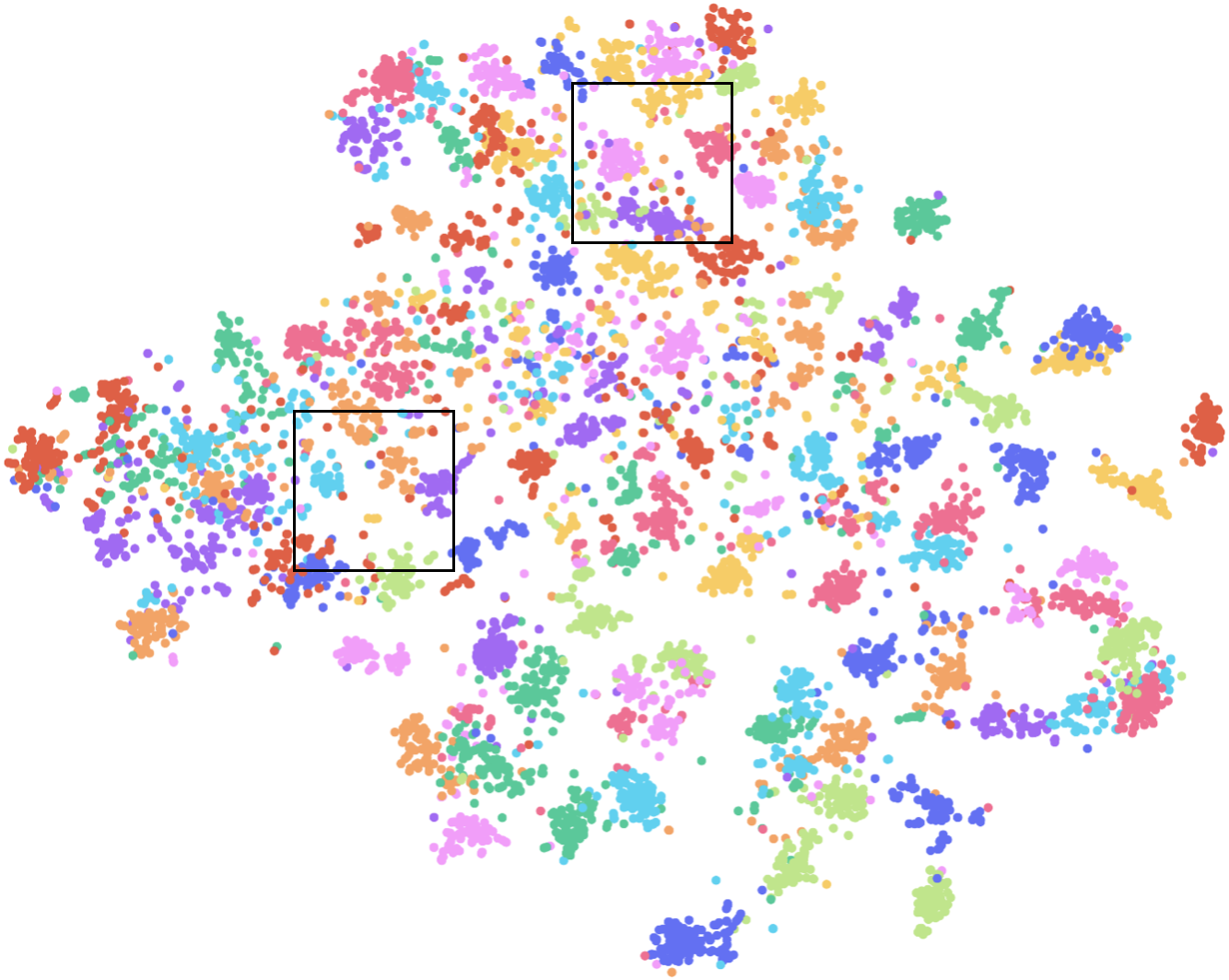}
        \caption{Test set, P-ISA}
        \label{fig4.1d}
    \end{subfigure}
    \caption{t-SNE visualization of the 512-dimensional embedding space for the Cars-196 dataset. (a) and (b) present the changes of the embedding space learned by Proxy-ISA during training at epoch 30 and 80, respectively, and (c) and (d) show the results of PA and P-ISA on the test set, respectively.}
    \label{fig4.1}
    \vspace{-3mm}
\end{figure}

\vspace{1mm}
\noindent
\textbf{Implementation Details} \hspace{2mm}
Our method was implemented in PyTorch \cite{NEURIPS2019_9015}.% , and all experiments were performed on NVIDIA A100 GPUs.
We used the Inception network with batch normalization \cite{pmlr-v37-ioffe15} as the embedding network, the embedding dimension was set to 512.
All input images were cropped to 224$\times$224, and random cropping and horizontal flipping were applied for augmentation during training; only center-cropping was used during testing.
The memory queue was turned on at the second epoch and the filter was enabled one epoch after that.
The hyper-parameters induced by Proxy-Anchor were set to $\alpha = 32$, $\delta = 10^{-1}$ following \cite{Kim_2020_CVPR} for all experiments.
Unless otherwise mentioned, we adopted a batch size of 128, and used the Adam optimizer \cite{DBLP:journals/corr/KingmaB14} with a learning rate of $10^{-4}$ and hyper-parameters $V=100$, $h = 1.5 \times 10^{-1}$, $k = 9 \times 10^{-1}$, $\lambda = 10^{-1}$, $\tau = 1.5$ as our default setting.

\subsection{Results}
%\subsection{Embedding Space Visualization}
\noindent
\textbf{Embedding Space Visualization} \hspace{2mm}
To outline how Proxy-ISA controls the learning of proxies, we visualize the embedding space via t-SNE \cite{vanDerMaaten2008}.
As illustrated in \cref{fig4.1a}, Proxy-ISA gradually widens the distance between proxies so that each proxy is distributed in its own subspace, and easier classes form clusters earlier than harder classes.
This demonstrates that the class hardness differs, and the definition of informative samples for each class also differs.
The relationships of class hardness learned by Proxy-ISA reserves a wider range for a harder class even if it is an unseen category, as illustrated in \cref{fig4.1d}.
This demonstrates that Proxy-ISA can better utilize the embedding space by learning dynamic margins according to the class hardness.
As a result, the model trained with Proxy-ISA acquires a more detailed subspace discriminative capability and thus a better generalization.

\setlength{\tabcolsep}{4pt}
\begin{table}[t]
    \centering
    \caption{Comparison of Recall@1 of proxy-based methods on four famous datasets for the task of image retrieval. $\dagger$ denotes evaluation in a fair setting}
    \begin{tabular}{@{}l c c c c@{}}
        \toprule
        Method & CUB & Cars & SOP & In-Shop \\
        \midrule
        Proxy-NCA \cite{8237309} & 65.0 & 83.2 & 75.9 & 86.9 \\
        SoftTriplet \cite{qian2019striple} & 65.4 & 84.5 & 78.3 & -- \\
        Proxy-Anchor$^\dagger$ \cite{Kim_2020_CVPR} & 66.6 & 84.0 & 78.3 & 91.3 \\
        ProxyGML \cite{Zhu2020ProxyGML} & 66.6 & 85.5 & 78.0 & -- \\
        Proxy-ISA (ours) & \textbf{68.1} & \textbf{86.4} & \textbf{78.9} & \textbf{92.3} \\
        \bottomrule
    \end{tabular}
    \label{table1}
    \vspace{-2mm}
\end{table}

\setlength{\tabcolsep}{2.2pt}
\begin{table}[t]
    \centering
    \caption{Evaluation in MLRC setting. $\dagger$ denotes evaluation in a fair setting}
    % \scalebox{0.78}{
    \begin{tabular}{@{}l | c c | c c@{}}
        \toprule
        \multirow{2}{4em}{Method} & \multicolumn{2}{c|}{CUB-200-2011} & \multicolumn{2}{c}{Cars-196} \\
        & R@1 & MAP@R & R@1 & MAP@R \\
        \midrule
        Margin \cite{8237571} & 63.6 $\pm$ 0.5 & 23.1 $\pm$ 0.3 & 81.2 $\pm$ 0.5 & 24.2 $\pm$ 0.3 \\
        Proxy-NCA \cite{8237309} & 65.0 $\pm$ 0.4 & 23.9 $\pm$ 0.3 & 83.6 $\pm$ 0.3 & 25.4 $\pm$ 0.3 \\
        CosFace \cite{DBLP:conf/cvpr/WangWZJGZL018} & 67.3 $\pm$ 0.3 & 26.7 $\pm$ 0.2 & 85.5 $\pm$ 0.2 & 27.6 $\pm$ 0.3 \\
        ArcFace \cite{Deng_2019_CVPR} & 67.5 $\pm$ 0.3 & 26.5 $\pm$ 0.2 & 85.4 $\pm$ 0.3 & 27.2 $\pm$ 0.3 \\
        MS \cite{wang2019multi} & 65.0 $\pm$ 0.3 & 24.7 $\pm$ 0.1 & 85.1 $\pm$ 0.3 & 28.1 $\pm$ 0.2 \\
        MS+Miner \cite{wang2019multi} & 67.7 $\pm$ 0.2 & 25.2 $\pm$ 0.2 & 83.7 $\pm$ 0.3 & 27.0 $\pm$ 0.4 \\
        SoftTriplet \cite{qian2019striple} & 66.2 $\pm$ 0.4 & 25.6 $\pm$ 0.2 & 84.5 $\pm$ 0.3 & 27.1 $\pm$ 0.2 \\
        Proxy-Anchor$^\dagger$ \cite{Kim_2020_CVPR} & 66.3 $\pm$ 0.3 & 25.7 $\pm$ 0.3 & 83.6 $\pm$ 0.4 & 27.1 $\pm$ 0.3 \\
        Proxy-ISA (ours) & \textbf{68.1 $\pm$ 0.3} & \textbf{26.8 $\pm$ 0.2} & \textbf{86.3 $\pm$ 0.2} & \textbf{29.3 $\pm$ 0.2} \\
        \bottomrule
    \end{tabular}
    % }
    \label{table3}
    \vspace{-2mm}
\end{table}

\vspace{1mm}
\noindent
\textbf{Comparison with State-of-the-Art} \hspace{2mm}
We now compare Proxy-ISA with state-of-the-art DML methods by performing image retrieval tasks.
The learning rates were set to $6 \times 10^{-4}$ for SOP and In-Shop.
We only adopted the warm-up epoch and the AdamW optimizer \cite{loshchilov2018decoupled} in the exceptional settings\footnote{See supplementary material for details. \label{note1}}
introduced by \cite{Kim_2020_CVPR} for these two datasets.
For a fair comparison, all such settings were also removed for Proxy-Anchor in experiments on CUB-200-2011 and Cars-196.
As shown in \Cref{table1}, the proposed Proxy-ISA achieves the best performance compared with the existing proxy-based methods.
% Proxy-ISA improves Recall@1 by 1.5\% on CUB-200-2011, 2.4\% on Cars-196, 0.6\% on SOP and 1\% on In-Shop compared with Proxy-Anchor.

The evaluation results in the MLRC setting are presented in \Cref{table3}.
As the MAP@R metric evaluates the whole embedding space, the results on the two widely used datasets demonstrate the effectiveness of Proxy-ISA in improving the quality of the embedding space.

\section{Conclusion}

Treating the proxy-data pairs equally to data pairs can cause undesirable proxy distribution in the embedding space.
To address this problem, we proposed the Informative Sample-Aware Proxy (Proxy-ISA), which controls gradient weights non-uniformly according to different semantic states.
Based on the adaptive semantic states for different classes, Proxy-ISA helps each proxy discover the most informative data by exploiting learned information.
Furthermore, we showed that the definition of ``informative" is class hardness dependent by applying the concept of ``\textit{volume}'' of the learned subspace to DML.
The empirical results demonstrate the superiority of Proxy-ISA over state-of-the-arts.% existing methods.

% \begin{acks}
% \end{acks}

% \section*{Acknowledgments}
\begin{acks}
    This work is supported by DENSO IT LAB Recognition and Learning Algorithm Collaborative Research Chair (Tokyo Tech.).
\end{acks}

\bibliographystyle{ACM-Reference-Format}
\bibliography{reference}


\end{document}

\title{Supplementary material for \textit{Informative Sample-Aware Proxy for Deep Metric Learning}}


\author{Aoyu Li}
 \email{aoyuli@rio.gsic.titech.ac.jp}
\affiliation{
  \institution{Tokyo Institute of Technology}
  %\city{Tokyo}
%  \country{Japan}
}

\author{Ikuro Sato}
%\email{isato@c.titech.ac.jp}
\affiliation{
  \institution{Tokyo Institute of Technology}
  \institution{Denso IT Laboratory}
%  \city{Tokyo}
%  \country{Japan}
}

\author{Kohta Ishikawa}
%\email{ishikawa.kohta@core.d-itlab.co.jp}
\affiliation{%
  \institution{Denso IT Laboratory}
%  \city{Tokyo}
%  \country{Japan}
}

\author{Rei Kawakami}
%\email{reikawa@sc.e.titech.ac.jp}
\affiliation{
  \institution{Tokyo Institute of Technology}
%  \city{Tokyo}
%  \country{Japan}
}

\author{Rio Yokota}
%\email{rioyokota@gsic.titech.ac.jp}
\affiliation{%
  \institution{Tokyo Institute of Technology}
%   \streetaddress{}
%  \city{Tokyo}
%   \state{}
%  \country{Japan}
%   \postcode{}
}

\renewcommand{\shortauthors}{Li, et al.}
% \renewcommand{\shortauthors}{Anonymized author, et al.}


\maketitle

\vspace{2mm}

\appendix

\setcounter{equation}{17}
\setcounter{figure}{6}
\setcounter{table}{2}

\section{Algorithm and Training Complexity}

We describe the training flow of our proposed method in \cref{alg1}.

\begin{algorithm}
    \caption{A training iteration of Proxy-ISA}\label{alg1}
    \begin{algorithmic}
        \State \textbf{Input:} data embeddings $\boldsymbol{X}$, class label $\boldsymbol{y}$, proxy set $\boldsymbol{P}$, memory queue $M$.
        % \Ensure $y = x^n$
        \State $\boldsymbol{S} \gets$ cosine similarity between $\boldsymbol{P}$ and $\boldsymbol{X}$
        \State $\boldsymbol{C}^+ \gets$ class labels appeared in $\boldsymbol{y}$
        \State $filter \gets \emptyset$
        \State $\eta_1, \eta_2, \dots, \eta_C \gets$ calculate search length for each class
        % \For {$c \in \boldsymbol{C}^+$}
        %     \State $freq_c \gets$ frequency of each class appeared in $\boldsymbol{X}$
        % \EndFor
        \For {$c \in \boldsymbol{C}^+$}
            \For {$i: y_i = c$}
                \If {$filter$ is turned on}
                    \State Calculate $\omega_{i, c}^+$ according to $\mathcal{S}_{i, c}$
                    \State Add $\boldsymbol{x}_i$ to $filter$ if $\mathcal{S}_{i, c} < \mathcal{S}_{learned}^{(c)} - \eta_c$
                \Else
                    \State $\omega_{i, c}^+ \gets 1$
                \EndIf
            \EndFor
        \EndFor
        \For {$c = 1, 2, \dots, C$}
            \For {$i: y_i \neq c$}
                \State Calculate $\omega_{ic}^-$ according to $\mathcal{S}_{i, c}$
            \EndFor
        \EndFor
        \State Calculate loss
        \If {$M$ is turned on}
            \State Enqueue $\boldsymbol{x}_i \in \{\boldsymbol{X} - filter\}$ to $M$
            \State Dequeue old embeddings if $M$ out of size
            \For {$c \in \boldsymbol{C}^+$}
                \State Update $\mathcal{S}_{learned}^{(c)}$ through $M$
            \EndFor
        \EndIf
    \end{algorithmic}
\end{algorithm}

\vspace{1mm}
% \subsubsection{Training Complexity}
\noindent \textbf{Training Complexity} \hspace{2mm}
Although our proposed Proxy-ISA takes complexity of $O(NC)$, it will increase the training time to some extent on large datasets that contain a larger number of categories (\ie, $C \gg N$), since each dynamic weight is treated independently and calculation of each $\eta_c$ takes extra complexity.

\section{Proof of Proposition 1}

\textit{Proof.} According to the K.K.T. conditions \cite{boyd2004convex}, $\mathcal{P}_c^+$ in %\cref{prop1} 
Eq.~(6) has the closed-form solution
\begin{equation}
    \mathcal{P}_c^+(i) = \frac{e^{\alpha(\delta - \mathcal{S}_{i, c})}}{1 + \underset{j: y_j = c}{\sum} e^{\alpha(\delta - \mathcal{S}_{j, c})}}.
\end{equation}
Therefore, we have
\begin{equation}
    \alpha \sum_{i: y_i = c} \mathcal{P}_c^+(i)(\delta - \mathcal{S}_{i, c}) + H(\mathcal{P}_c^+) = \log\left( 1 + \sum_{i: y_i = c} e^{\alpha(\delta - \mathcal{S}_{i, c})} \right).
\end{equation}
The same analysis is applicable to $\mathcal{P}_c^-$.

\begin{figure*}
    \centering
    \begin{subfigure}[b]{0.32\linewidth}
        \centering
        \includegraphics[width=\linewidth]{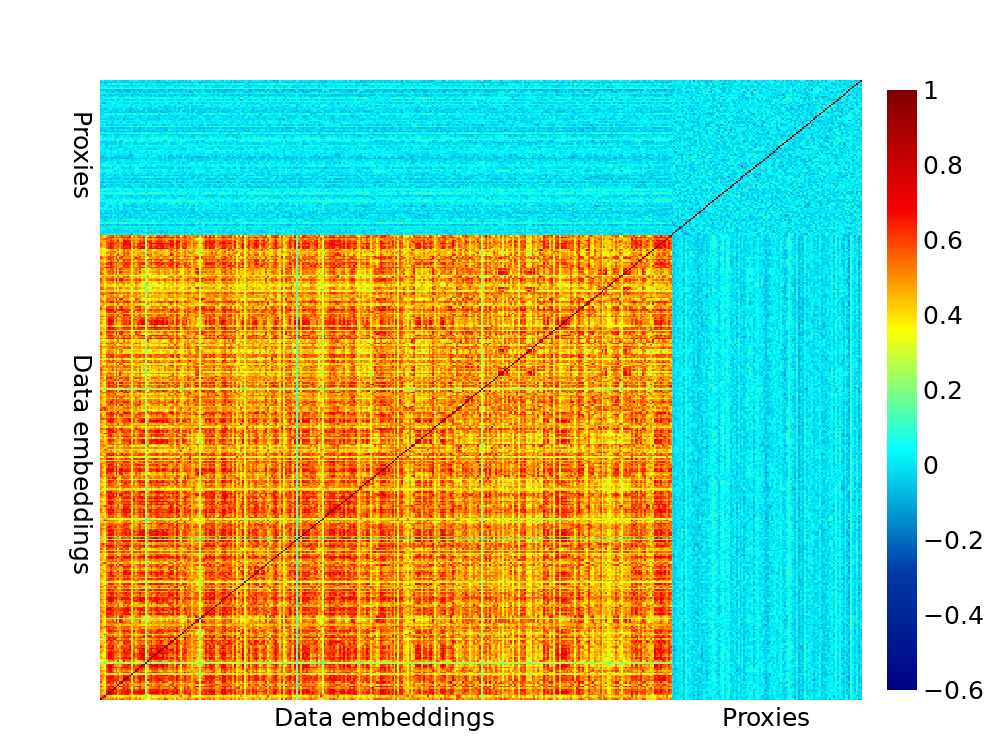}
        \caption{N.SoftMax@Initial}
        \label{supFig4a}
    \end{subfigure}
    \hfill
    \begin{subfigure}[b]{0.32\linewidth}
        \centering
        \includegraphics[width=\linewidth]{img/HeatMapInit.png}
        \caption{Proxy-Anchor@Initial}
        \label{supFig4b}
    \end{subfigure}
    \hfill
    \begin{subfigure}[b]{0.32\linewidth}
        \centering
        \includegraphics[width=\linewidth]{img/HeatMapInit.png}
        \caption{Proxy-ISA@Initial}
        \label{supFig4c}
    \end{subfigure}
    \begin{subfigure}[b]{0.32\linewidth}
        \centering
        \includegraphics[width=\linewidth]{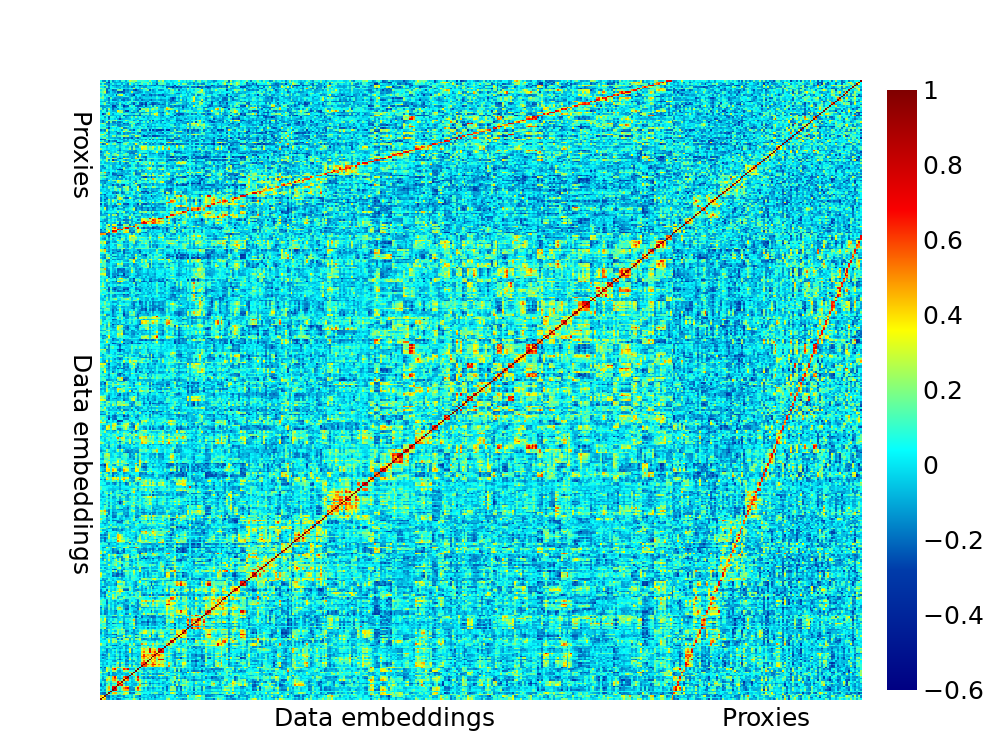}
        \caption{N.SoftMax@eph10}
        \label{supFig4d}
    \end{subfigure}
    \hfill
    \begin{subfigure}[b]{0.32\linewidth}
        \centering
        \includegraphics[width=\linewidth]{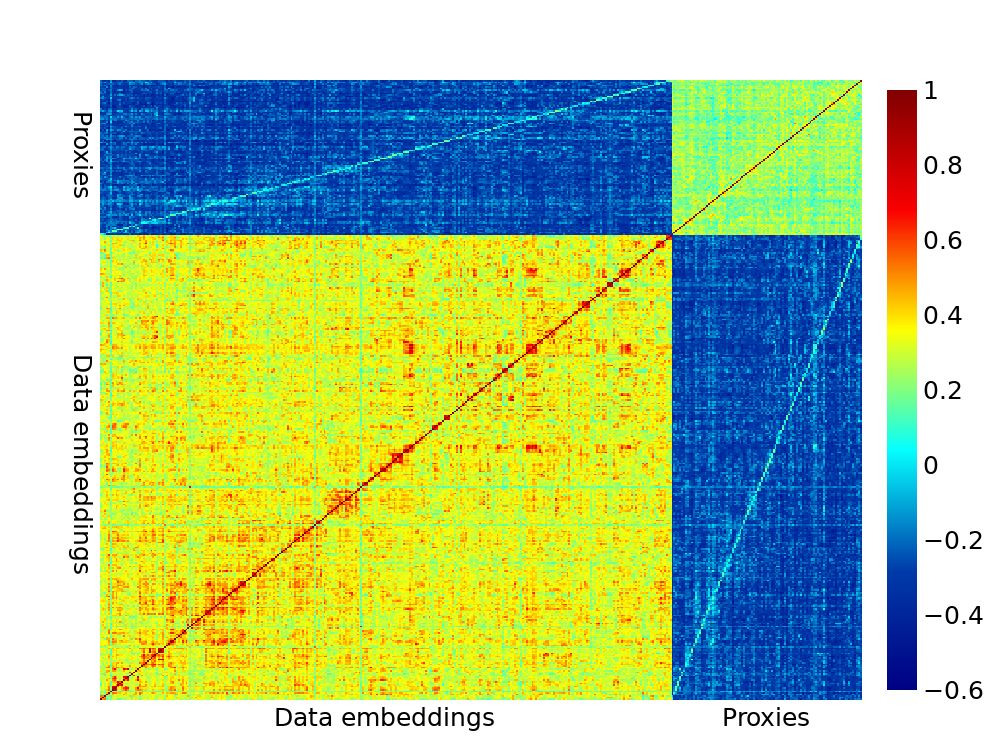}
        \caption{Proxy-Anchor@eph10}
        \label{supFig4e}
    \end{subfigure}
    \hfill
    \begin{subfigure}[b]{0.32\linewidth}
        \centering
        \includegraphics[width=\linewidth]{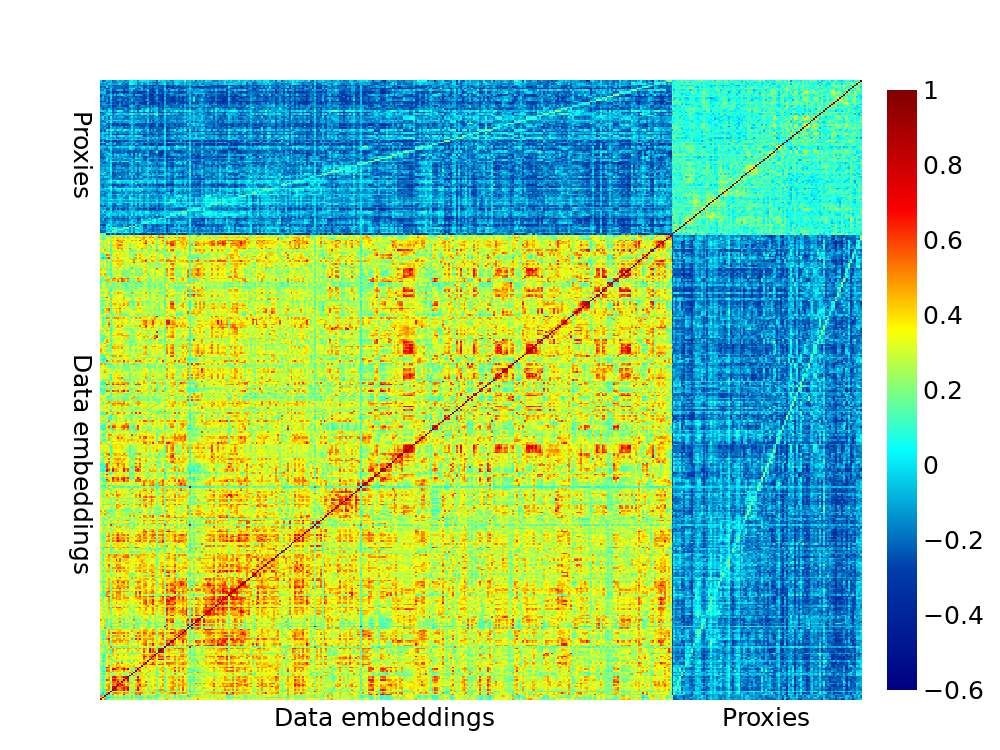}
        \caption{Proxy-ISA@eph10}
        \label{supFig4f}
    \end{subfigure}
    \begin{subfigure}[b]{0.32\linewidth}
        \centering
        \includegraphics[width=\linewidth]{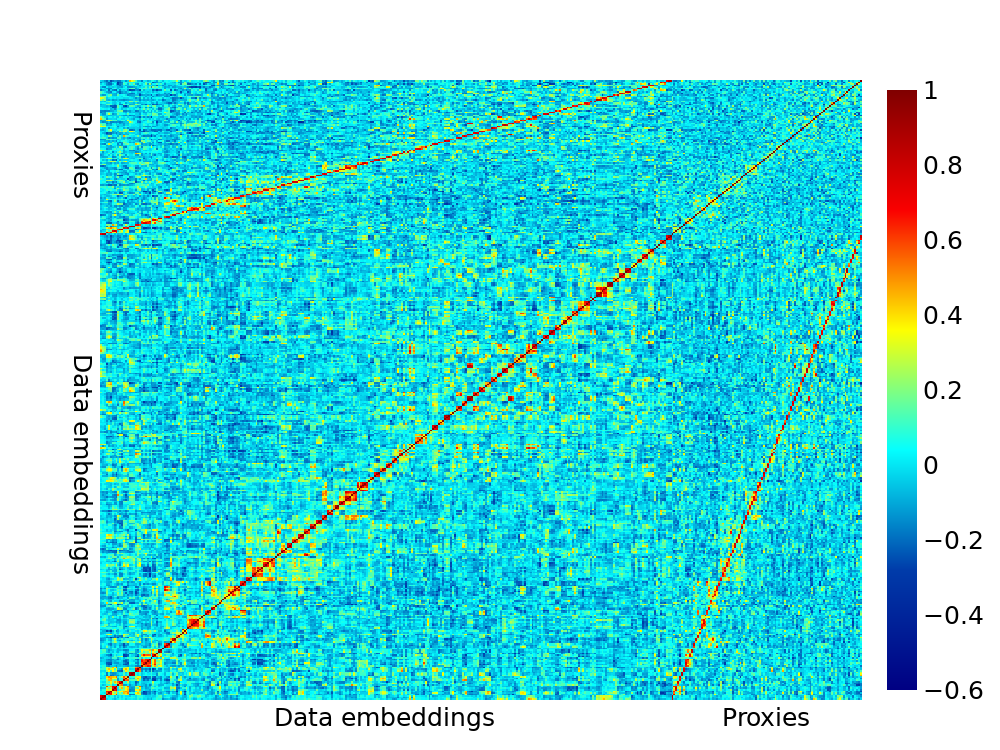}
        \caption{N.SoftMax@eph30}
        \label{supFig4g}
    \end{subfigure}
    \hfill
    \begin{subfigure}[b]{0.32\linewidth}
        \centering
        \includegraphics[width=\linewidth]{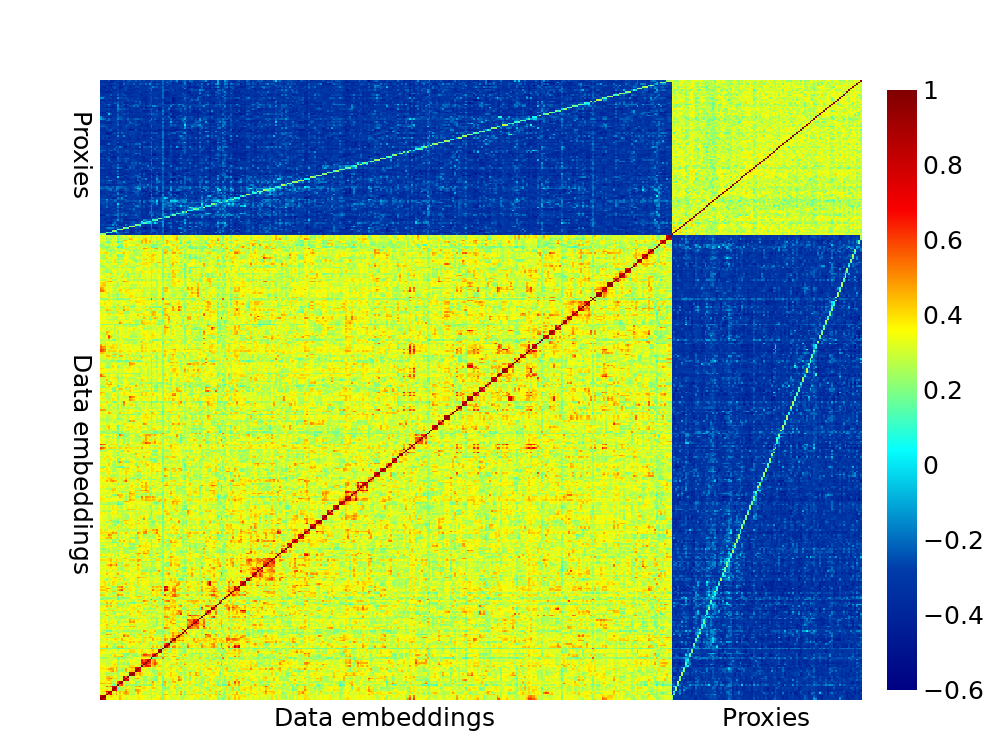}
        \caption{Proxy-Anchor@eph30}
        \label{supFig4h}
    \end{subfigure}
    \hfill
    \begin{subfigure}[b]{0.32\linewidth}
        \centering
        \includegraphics[width=\linewidth]{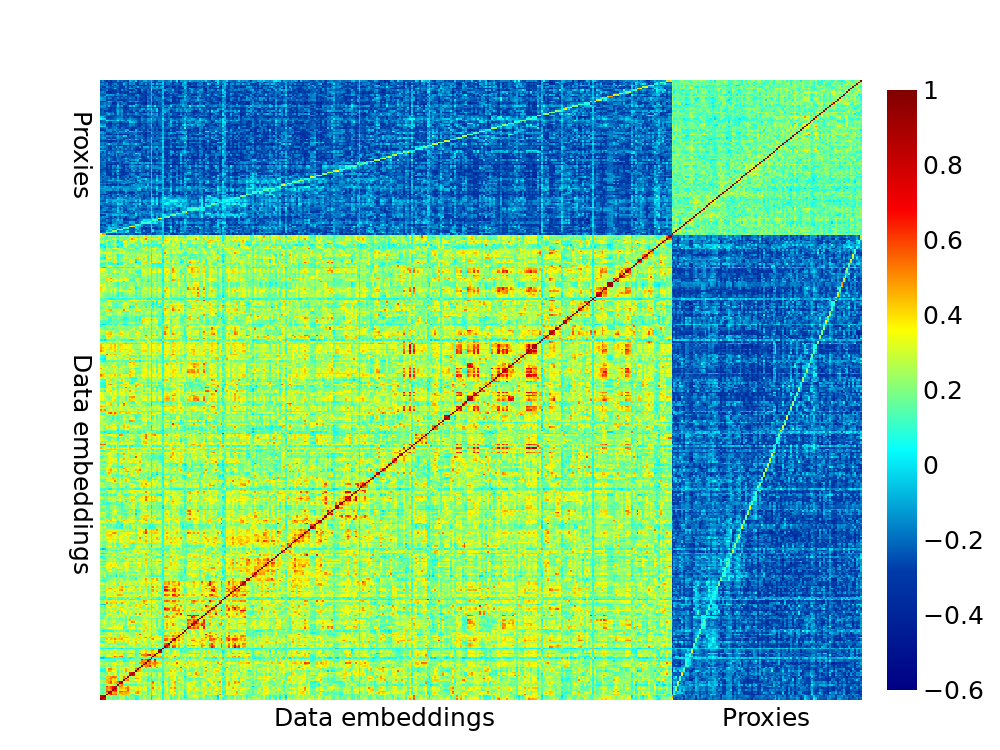}
        \caption{Proxy-ISA@eph30}
        \label{supFig4i}
    \end{subfigure}
    \begin{subfigure}[b]{0.32\linewidth}
        \centering
        \includegraphics[width=\linewidth]{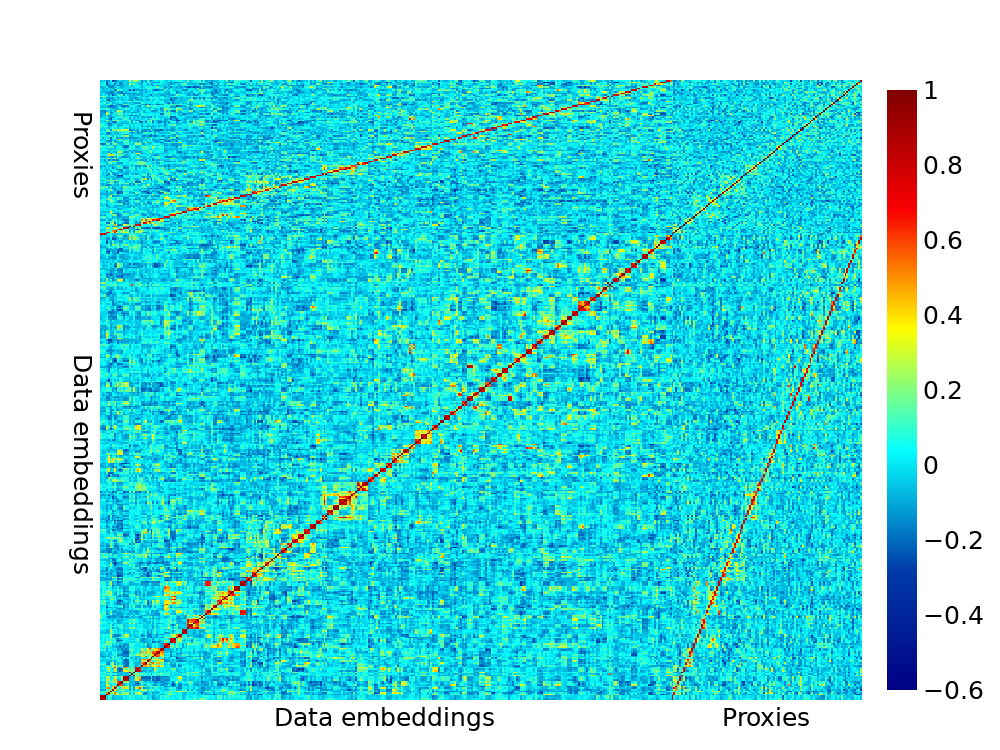}
        \caption{N.SoftMax@eph60}
        \label{supFig4j}
    \end{subfigure}
    \hfill
    \begin{subfigure}[b]{0.32\linewidth}
        \centering
        \includegraphics[width=\linewidth]{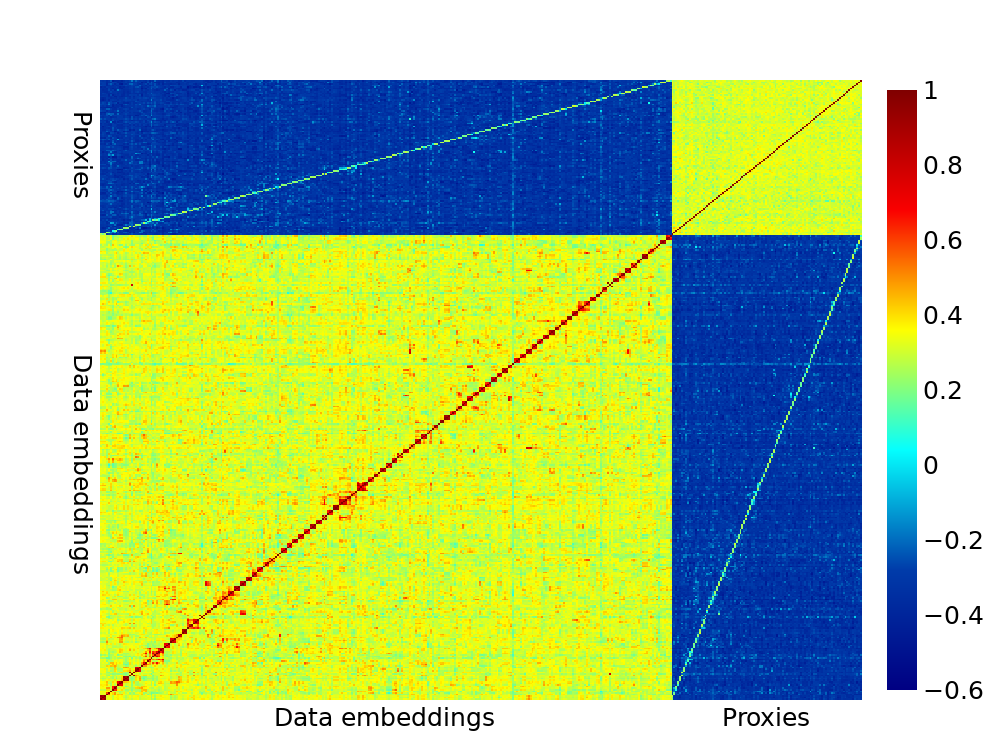}
        \caption{Proxy-Anchor@eph60}
        \label{supFig4k}
    \end{subfigure}
    \hfill
    \begin{subfigure}[b]{0.32\linewidth}
        \centering
        \includegraphics[width=\linewidth]{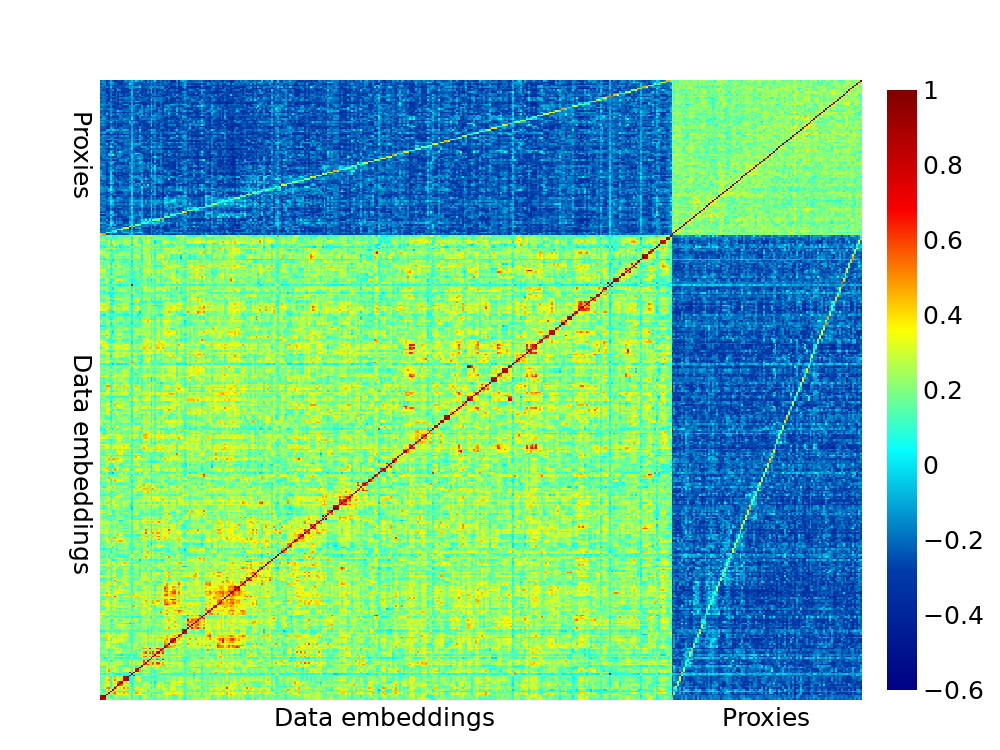}
        \caption{Proxy-ISA@eph60}
        \label{supFig4l}
    \end{subfigure}
    \caption{Heat map of cosine similarity between data embeddings of training set of the Cars-196 dataset and proxies. Three data samples are randomly selected for each class.}
    \label{supFig4}
\end{figure*}
% \clearpage

\section{Details of the Decay Function}

Since the penalty from $\omega_{i, c}^+$ is decided by the estimated class hardness, it should not be too large when the model cannot determine the class hardness.
In other words, all $\omega_{i, c}^+$ should be close to $1$, with less variation in the early learning phase.
Given an original sigmoid function that varies with $E_n^{(c)}$:
\begin{equation}
    \frac{1}{1 + e^{-E_n^{(c)}}},
\end{equation}
we then adjust it to our desired curve.
First, to satisfy $\lim_{n \to 0+} \sigma_n^{(c)} \approx 1$, we transform the function to
\begin{equation}
    1 - \frac{1}{1 + e^{-(E_n^{(c)} - V) - \tau}},
\end{equation}
where we introduce $\tau$ to control the timing of decay.
Secondly, to satisfy $\lim_{n \to \infty} \sigma_n^{(c)} = \nu_n^{(c)}$, we formulate an equation with unknown variable $\xi$ as follows:
\begin{equation}
    \lim_{n \to \infty} \sigma_n^{(c)} = 1 - \frac{\xi}{1 + e^{-\tau}} = \nu_n^{(c)}.
\end{equation}
The solution is
\begin{equation}
    \xi = (1 + e^{-\tau})(1 - \nu_n^{(c)}),
\end{equation}
thus, the decay function is formulated as:
\begin{equation}
    \sigma_n^{(c)} = 1 + \frac{(1 + e^{-\tau})(\nu_n^{(c)} - 1)}{1 + e^{V - E_n^{(c)} - \tau}}.
\end{equation}

The penalty from $\omega_{i, c}^-$ is set to $1 / E_n^{(c)}$ ($< \nu_n^{(c)}$) for preventing the repulsive force on the proxy from being too large under the impact of a large population of easy negatives.

\begin{figure}[t]
    \centering
    \begin{subfigure}[b]{0.41\linewidth}
        \centering
        \includegraphics[width=\linewidth]{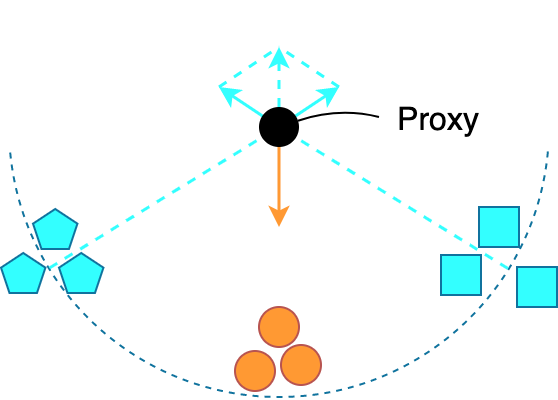}
        \caption{A loose space}
        \label{supFig3a}
    \end{subfigure}
    \hfill
    \begin{subfigure}[b]{0.4\linewidth}
        \centering
        \includegraphics[width=\linewidth]{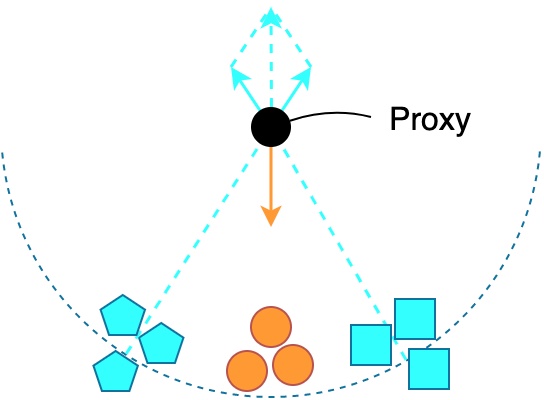}
        \caption{A compact space}
        \label{supFig3b}
    \end{subfigure}
    \caption{Impact of compactness of the embedding space on the repulsive force. A compact space (or clusters of the harder classes) generates greater joint repulsive force, even if the cosine similarity to the proxy (\ie\ the gradient weight of the proxy-data pairs) is kept unchanged.}
    \label{supFig3}
\end{figure}
% \clearpage

\section{Impact of Proxy-ISA on the Embedding Space}

To quantitatively describe the learned embedding space, we show its compactness via heat map of pair-wise cosine similarity.
As illustrated in \cref{supFig4}, the distribution of the data embeddings is relatively compact at the initial stage (results of the pre-trained backbone and a randomly initialized embedding layer), and is dispersed to varying degrees after applying different objective functions.
\cref{supFig4e}, \cref{supFig4h} and \cref{supFig4k} show that the embedding space learned by Proxy-Anchor loss \cite{Kim_2020_CVPR} stops dispersing after it reaches a certain stage, compared to the case of normalized SoftMax loss in \cref{supFig4d}, \cref{supFig4g} and \cref{supFig4j}.
This is resulted from the key difference between Proxy-Anchor and normalized SoftMax, the consideration of the optimization problem (\ie, Proxy-Anchor only considers the optimal distribution of data embeddings related to each proxy, while normalized SoftMax only considers the optimal distribution of proxies related to each data embedding), since Proxy-Anchor generates stronger attractive and repulsive force between a proxy and a relatively hard sample, the proxy distribution is easily disrupted.

Since most of the data embeddings are scattered in a relatively compact space at the initial stage, a larger fraction of this is regarded as hard samples (mostly negative) by Proxy-Anchor, which means the proxies are more likely to move in undesired directions, thus forming a cluster far from the data embeddings, as shown in the paper.
After the proxies reach a balance (\ie, learned a shortcut by discriminating the data embeddings while getting far from them), they still keep the ability to force the positives to be close to each other and to be far from the negatives, but failed to disperse the whole embedding space further.
In the meanwhile, a too compact embedding space generates stronger repulsive force, which blocks the proxy from getting closer to its ideal cluster, as described in \cref{supFig3}.

Unlike Proxy-Anchor, our proposed Proxy-ISA captures informative samples that correspond to different learning stages, thus continues to disperse the embedding space, even in the late learning phase, as shown in \cref{supFig4i} and \cref{supFig4l}.
Although the dispersion is not as great as normalized SoftMax, Proxy-ISA allows the model to consider relative sample hardness for each class, which is ignored by normalized SoftMax (also ignored by other SoftMax based methods \cite{DBLP:conf/cvpr/WangWZJGZL018,Deng_2019_CVPR} and Proxy-NCA \cite{8237309}).
In other words, Proxy-ISA considers both sample-wise hardness and class-wise hardness, this helps the proxy discover its own informative samples adaptively according to different classes, and thus generates better learning signals for the model.
% This demonstrates that Proxy-ISA can better utilize the embedding space by learning dynamic margins according to the class hardness.

\section{Ablation Study}

\subsection{Impact of Hyper-parameters}

\begin{figure}[t]
    \centering
    \includegraphics[width=0.7\linewidth]{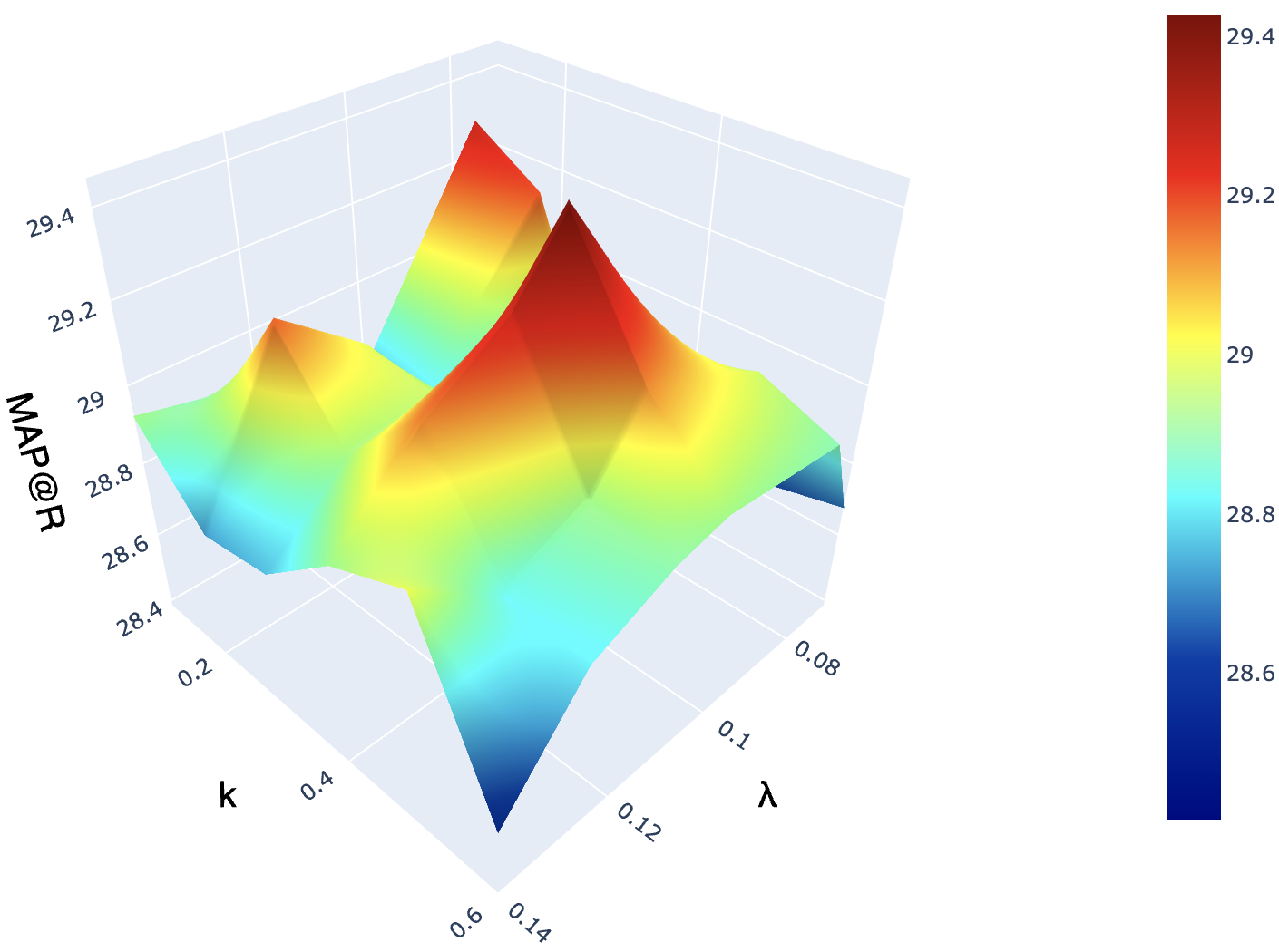}
    \caption{Impact of the hyper-parameters $k$ and $\lambda$ on the MAP@R metric.}
    \label{supFig5}
\end{figure}

The most important hyper-parameters are $k$ and $\lambda$, which control the search length ($\eta_c$) of the informative samples, and $k$ also controls the sensitivity to class hardness.
As illustrated in \cref{supFig5}, for Cars-196 \cite{6755945} dataset, when $k$ reaches an appropriate range, the adaptive search length helps each proxy focus on its informative samples, and thus improves the model performance for a wider range of the margin $\lambda$.
In the meanwhile, a search length with too large a $k$, \ie, too sensitive to the class hardness, will be more likely to introduce outliers for harder classes, thus can degrade the generalization.
This is consistent with that hard samples do not always mean informative.
For different datasets, the appropriate $k$ should be different because the inter-class relation differs.

% \subsection{Impact of $V$ and Memory Size $T$}

\begin{figure}[t]
    \centering
    \begin{subfigure}[b]{0.47\linewidth}
        \centering
        \includegraphics[width=\linewidth]{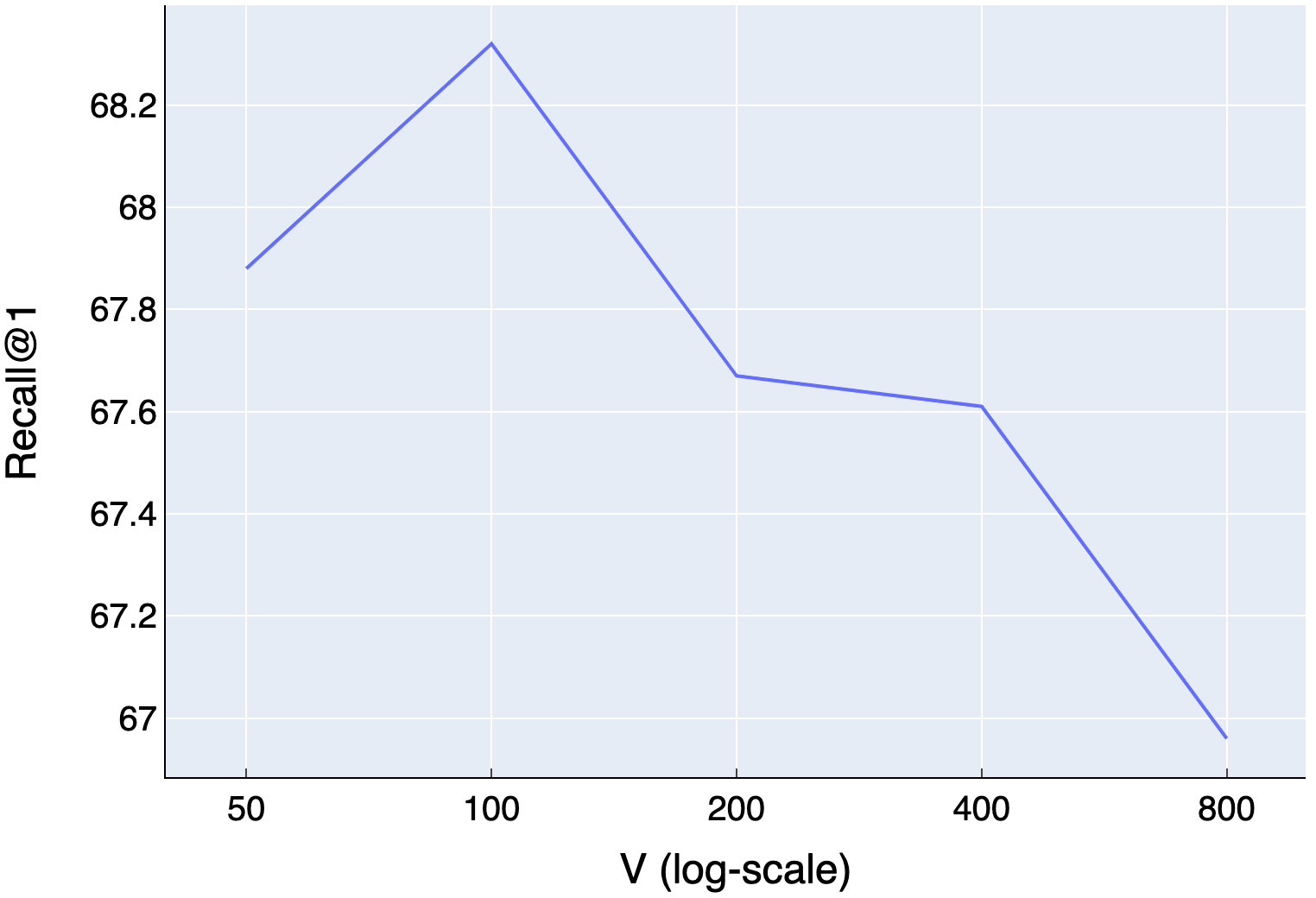}
        \caption{Recall@1 results \vs\ $V$}
        \label{supFig1a}
    \end{subfigure}
    \hfill
    \begin{subfigure}[b]{0.47\linewidth}
        \centering
        \includegraphics[width=\linewidth]{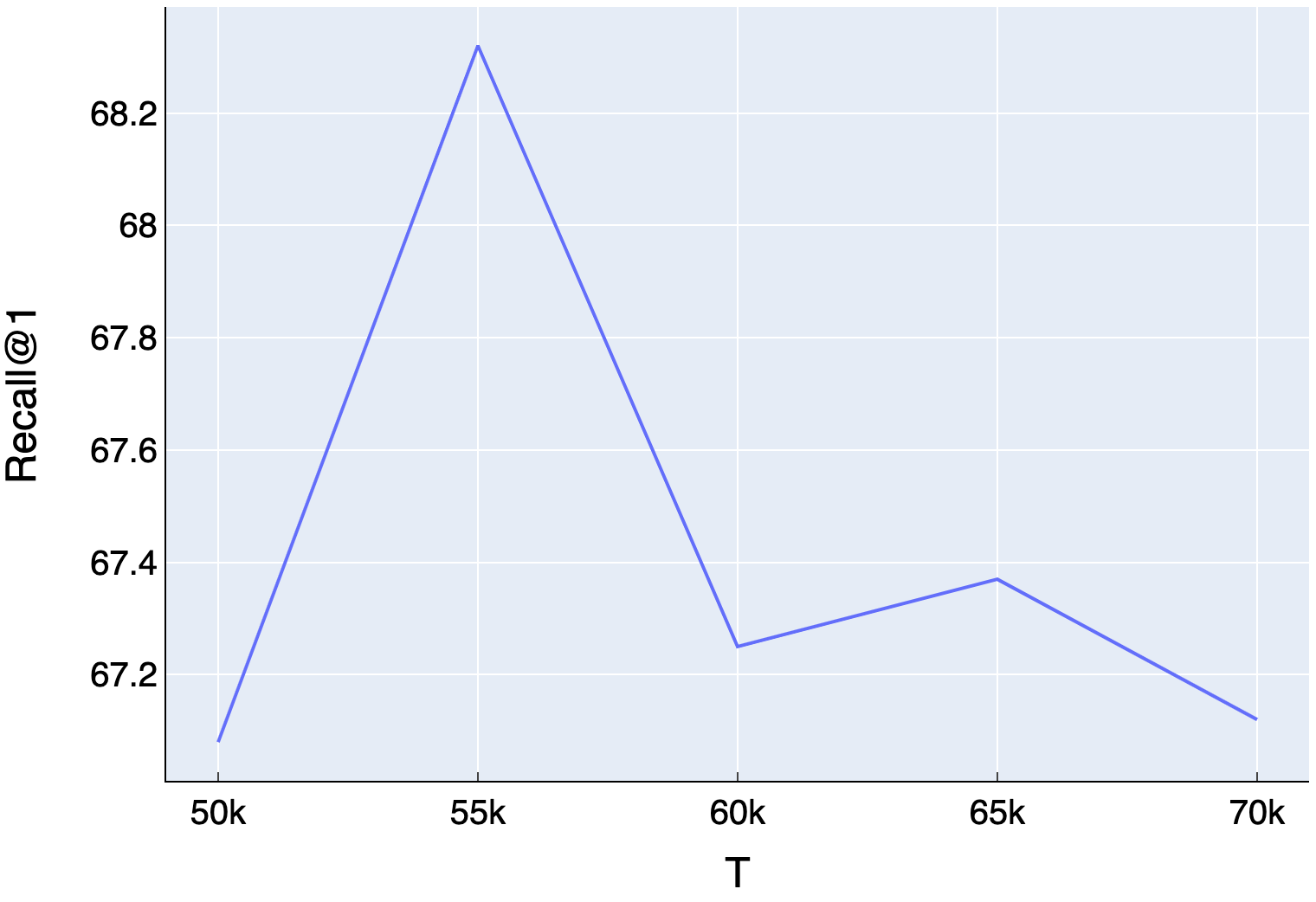}
        \caption{Recall@1 results \vs\ $T$}
        \label{supFig1b}
    \end{subfigure}
    \caption{Impact of the hyper-parameters $V$ (upper bound of $E_n$) and $T$ (maximum size of the memory queue) on the Recall@1 metric.}
    \label{supFig1}
\end{figure}

We show the impact of $V$ and $T$ on the test set of CUB-200-2011 \cite{WahCUB_200_2011} (100 classes).
As illustrated in \cref{supFig1a}, too large $V$ will degrade the model performance, since total volume of the finite samples in the feature space of one class has an appropriate limit, which should not be set too large.
In the meanwhile, $V$ controls the decay of the dynamic weights, which needs to be at an appropriate timing to help the model focus on the informative samples.

The memory size $T$ affects the estimation quality of $\mathcal{S}_{learned}^{(c)}$, which is the threshold of a learned subspace.
Too small $T$ will result in an insufficient number of samples used for estimation, while too large $T$ will result in some of the samples in $M$ being outdated, thus the memory size has a best range, as shown in \cref{supFig1b}.

\subsection{Effect of the Decay Function}

To show the effect of the decay function $\sigma_n^{(c)}$, we compare the performance by setting different penalties of $\omega_{i, c}^+$.
As shown in \Cref{supTable2}, setting $\omega_{i, c}^+ = 1 / E_n^{(c)}$, which is the same to $\omega_{i, c}^-$, will degrade the performance since the imbalanced weighting between positives and negatives, as discussed in the paper, bounding it with $\nu_n^{(c)}$ alleviates this problem.
After applied the decay function $\sigma_n^{(c)}$ for $\omega_{i, c}^+$, our method provides weighting factors dynamically along learning, and the penalty increases significantly only when the learning of the class reached an appropriate phase, this also helps the model treat informative samples non-uniformly along learning.

\begin{table}[t]
    \centering
    \caption{Comparison of different combination of penalty from $\omega_{i, c}$ on Cars-196 dataset}
    \begin{tabular}{@{}l c c c c c@{}}
        \toprule
        $\omega_{i, c}^+$: & $1$ (no penalty) & $1$ (no penalty) & $1 / E_n^{(c)}$ & $\nu_n^{(c)}$ & $\sigma_n^{(c)}$ \\
        $\omega_{i, c}^-$: & $1$ (no penalty) & $1 / E_n^{(c)}$ & $1 / E_n^{(c)}$ & $1 / E_n^{(c)}$ & $1 / E_n^{(c)}$ \\
        \midrule
        Recall@$1$ & 83.6 & 85.1 & 71.4 & 85.8 & \textbf{86.3} \\
        MAP@$R$ & 27.1 & 28.3 & 16.3 & 28.7 & \textbf{29.3} \\
        \bottomrule
    \end{tabular}
    \label{supTable2}
\end{table}

\begin{figure}[t]
    \centering
    \begin{subfigure}[b]{0.47\linewidth}
        \centering
        \includegraphics[width=\linewidth]{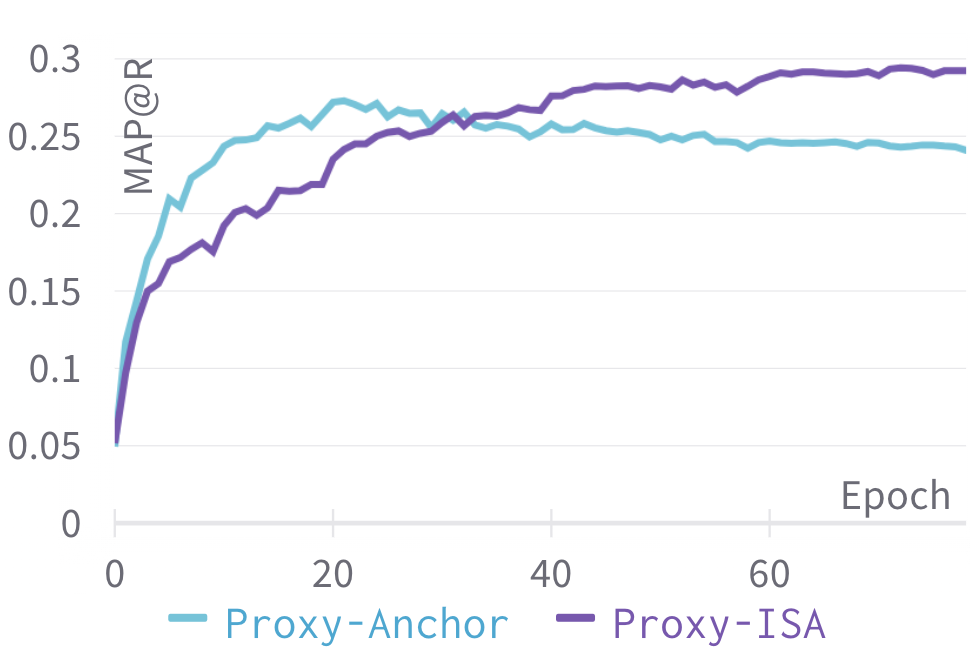}
        \caption{}
        \label{supFig2a}
    \end{subfigure}
    \hfill
    \begin{subfigure}[b]{0.47\linewidth}
        \centering
        \includegraphics[width=\linewidth]{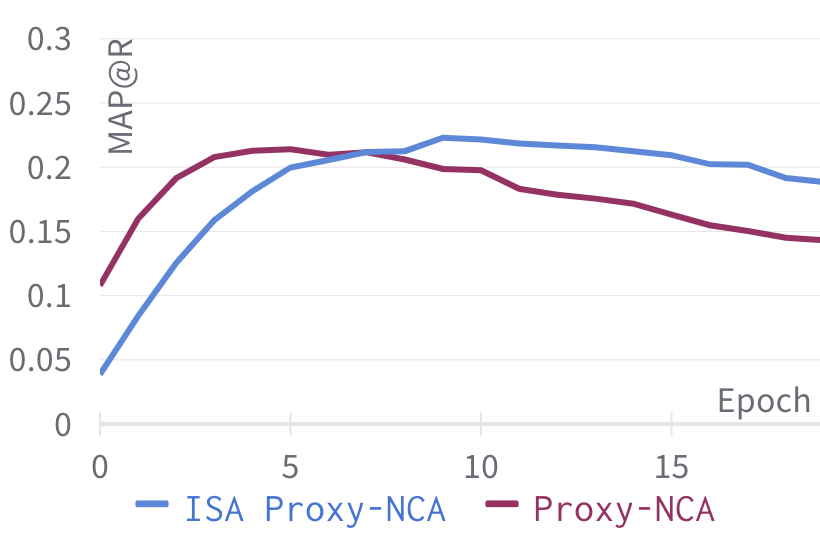}
        \caption{}
        \label{supFig2b}
    \end{subfigure}
    \caption{Performance on unseen classes of Cars-196. Although the baselines converge faster, over-fitting happens as learning progresses.}
    \label{supFig2}
\end{figure}

In the meanwhile, it prevents the gradient from being dominated by a relatively large number of easy proxy-data pairs by directly reducing their gradient weights, which also keeps the class-related regions from being over-compressed, thus prevents over-fitting, as shown in \cref{supFig2a}.
A similar effect can be observed when we apply the adaptive weighting scores directly on Proxy-NCA \cite{8237309} (\cref{supFig2b}).

\subsection{Extra Settings for Fair Comparison}

As mentioned in other work \cite{Zhu2020ProxyGML,ko2021learning}, Proxy-Anchor loss is actually implemented with three additional tricks: 1) the AdamW \cite{loshchilov2018decoupled} optimizer instead of Adam \cite{DBLP:journals/corr/KingmaB14}, 2) a parameter warm-up strategy for the last FC layer, 3) a combination of average and max pooling following the backbone network.
Comparison to Proxy-Anchor with all the exceptional settings enabled is shown in \Cref{supTable1}.

\begin{table}[t]
    \centering
    \caption{Recall@1 compared to Proxy-Anchor with exceptional settings applied}
    \begin{tabular}{@{}l c c c c@{}}
        \toprule
        Method & CUB & Cars & SOP & In-Shop \\
        \midrule
        Proxy-Anchor \cite{Kim_2020_CVPR} & 68.4 & 86.1 & 79.1 & 91.5 \\
        Proxy-ISA & \textbf{69.4} & \textbf{86.8} & \textbf{79.3} & \textbf{92.5} \\
        \bottomrule
    \end{tabular}
    \label{supTable1}
\end{table}

\bibliographystyle{ACM-Reference-Format}
\bibliography{reference}